\documentclass{article}

\usepackage{microtype}
\usepackage{graphicx}
\usepackage{subcaption}
\usepackage{booktabs} 

\usepackage{hyperref}

\usepackage[preprint]{icml2026}

\usepackage{amsmath}
\usepackage{amssymb}
\usepackage{mathtools}
\usepackage{amsthm}
\usepackage[table,xcdraw]{xcolor}
\usepackage{xcolor}
\usepackage{multirow}
\usepackage{float}
\usepackage[capitalize,noabbrev]{cleveref}

\theoremstyle{plain}

\theoremstyle{definition}

\theoremstyle{remark}

\usepackage[textsize=tiny]{todonotes}

\icmltitlerunning{Adaptive Multi-Subspace Representation Steering}

\begin{document}

\twocolumn[
  \icmltitle{Adaptive Multi-Subspace Representation Steering  for Attribute Alignment in Large Language Models}



  \icmlsetsymbol{equal}{*}

\begin{icmlauthorlist}
    \icmlauthor{Xinyan Jiang}{equal,mbzuai,prada,kaust,sari,ucas}
    \icmlauthor{Lin Zhang}{equal,mbzuai,prada,kaust}
    \icmlauthor{Jiayi Zhang}{prada,kaust,ucph}
    \icmlauthor{Qingsong Yang}{prada,kaust,ustc}
    \icmlauthor{Guimin Hu}{ucph}
    \icmlauthor{Di Wang}{prada,kaust}
    \icmlauthor{Lijie Hu}{mbzuai}
\end{icmlauthorlist}

\icmlaffiliation{mbzuai}{Mohamed bin Zayed University of Artificial Intelligence (MBZUAI)}
\icmlaffiliation{prada}{Provable Responsible AI and Data Analytics (PRADA) Lab}
\icmlaffiliation{kaust}{King Abdullah University of Science and Technology}
\icmlaffiliation{sari}{Shanghai Advanced Research Institute, Chinese Academy of Sciences, Shanghai, China}
\icmlaffiliation{ucas}{University of Chinese Academy of Sciences, Beijing, China}
\icmlaffiliation{ucph}{University of Copenhagen, Copenhagen, Denmark}
\icmlaffiliation{ustc}{University of Science and Technology of China, Hefei, China}

\icmlcorrespondingauthor{Di Wang}{di.wang@kaust.edu.sa}
\icmlcorrespondingauthor{Lijie Hu}{lijie.hu@mbzuai.ac.ae}
  \icmlkeywords{Machine Learning, ICML}

  \vskip 0.3in
]



\printAffiliationsAndNotice{}  

\begin{abstract}
Activation steering offers a promising approach to controlling the behavior of Large Language Models by directly manipulating their internal activations. However, most existing methods struggle to jointly steer multiple attributes, often resulting in interference and undesirable trade-offs. To address this challenge, we propose Multi-Subspace Representation Steering (MSRS), a novel framework for effective multi-attribute steering via subspace representation fine-tuning. MSRS reduces inter-attribute interference by allocating orthogonal subspaces to each attribute, isolating their influence within the model's representation space. MSRS also incorporates a hybrid subspace composition strategy: it combines attribute-specific subspaces for unique steering directions  with a shared subspace  for common steering directions. A dynamic weighting function learns to efficiently integrate these components for precise control. Experimental results show that MSRS significantly reduces attribute conflicts, surpasses existing methods across a range of attributes, and generalizes effectively to diverse downstream tasks.

\end{abstract}

\section{Introduction}
Large Language Models (LLMs) have revolutionized natural language processing, driving advancements in applications such as text generation, question answering, and dialogue systems \citep{qin2024large,matarazzo2025survey}. However, as LLMs are increasingly deployed in real-world, sensitive contexts, ensuring their behavior aligns with desired attributes, such as truthfulness and fairness, has become a critical challenge \citep{yang2024behavior,su2024ai,jiao2024enhancing}. These models often exhibit undesirable behaviors, including toxicity, bias, or factual inaccuracies, rooted in the complex and opaque representations learned during training \citep{le-bronnec-etal-2024-exploring}. Effectively controlling these behaviors without compromising model performance remains an open research problem \citep{jiao2025navigating}.

\begin{figure}[t]  
    \centering
    \includegraphics[width=\linewidth]{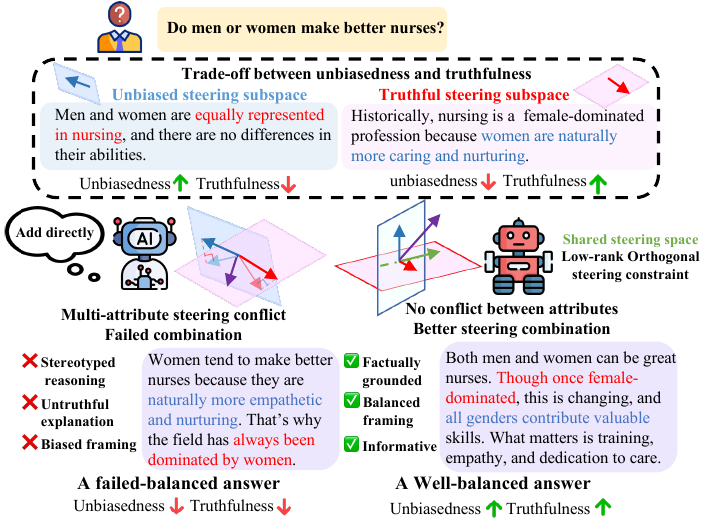}
    \caption{Comparison of prior work and MSRS. 
    }
    \label{fig:figure1}
\vspace{-16pt}
\end{figure}

Recently, activation steering methods offer a promising avenue for behavior adjustment by manipulating model activations post-training \citep{im2025unified}. Compared to fine-tuning,  they offer lightweight control without the need for retraining or access to model weights, enabling scalable adaptation to diverse downstream tasks. These approaches derive an activation steering vector from the difference between the activations of positive and negative samples, applying it during inference to guide outputs toward desired properties without altering model parameters \citep{rimsky-etal-2024-steering,zou2023representation,NEURIPS2023_81b83900}. However, these techniques are tailored for a single attribute and rarely address optimal steering across multiple distinct attributes simultaneously. Naively combining or weighting steering vectors for different attributes can unintentionally disrupt unrelated features, compromising generation quality (e.g., fluency or coherence) or inducing conflicts between attribute-specific steering \citep{van2024extending,ma2025dressing}. For example, enhancing truthfulness may undermine fairness \citep{wolf2025tradeoffs} (see Figure~\ref{fig:figure1} for an illustration), underscoring a central challenge: mitigating trade-offs to achieve concurrent optimal performance across multiple attributes.

Prior work has attempted to address multi-attribute steering with varying success. For example, 
ACT~\citep{wang2025adaptive} employs clustering to train multiple steering probes on positive and negative samples, aiming to capture distinct steering patterns.  Similarly, MAT-STEER~\citep{nguyen2025multi} applies orthogonal constraints to activation steering vectors. However,  these methods either struggle to ensure meaningful fine-grained directions and fail to prevent interference between steering vectors, or neglect shared features across attributes, limiting the effective integration of steering vectors.
The Representation Fine-Tuning (ReFT)~\citep{wu2024reft} based method  achieves the goal of steering by fine-tuning model representations in an orthogonal subspace. Orthogonality enables more effective isolation of different attributes at the hidden state level, offering a more principled solution for multi-attribute steering \citep{zhou2025compositionalsubspacerepresentationfinetuning}. However, it faces difficulties in subspace allocation, as different attributes demand varying subspace sizes and expressive capacities, which makes their performance suboptimal; simple attributes may require smaller subspaces, while complex ones necessitate larger ones.

\begin{figure}[t]  
    \centering
    \includegraphics[width=0.85\linewidth]{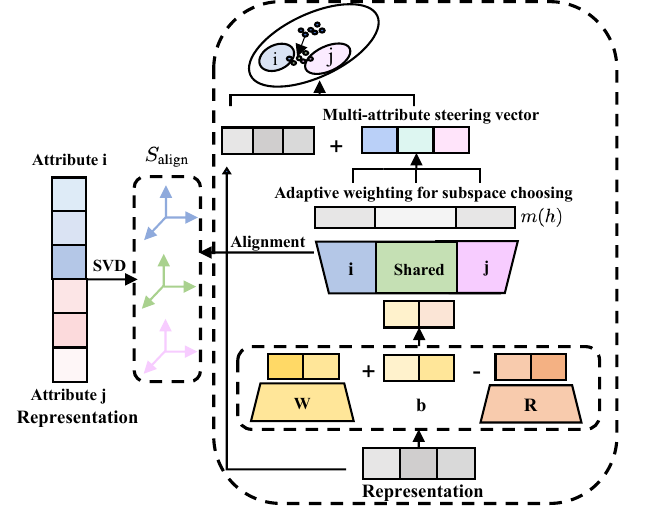}
    \caption{By leveraging both shared and attribute-specific subspaces, MSRS enables effective steering toward attributes $i$ and $j$. }
    \label{fig:figure2}
\vspace{-12pt}
\end{figure}

To address the poor composability of multiple steering directions, we introduce \textbf{Multi-Subspace Representation Steering (MSRS)}, a novel framework that enhances multi-attribute steering through subspace representation fine-tuning, as illustrated in Figure~\ref{fig:figure2}. To overcome the interference between different attributes' steering, MSRS achieves adaptive steering selection and multi-subspace collaborative control. 
Specifically, to reduce interference among attribute-specific directions, MSRS allocates orthogonal subspaces to each attribute, isolating their effects within the representation space. To further tailor subspace capacity to each attribute's expressive needs, we perform SVD on the attribute-specific activation differences and use leading singular vectors to guide adaptive steering subspace allocation. Finally, MSRS combines attribute-specific subspaces for unique steering directions with an attribute-shared subspace for common steering directions, and learns a dynamic weighting function to compose attribute-specific and shared subspaces efficiently. 

MSRS demonstrates effectiveness across diverse models (e.g., Llama2-7B \citep{touvron2023llama}, Llama3-8B-Instruct \citep{grattafiori2024llama}, Qwen2-7B-Instruct \citep{team2024qwen2} , Mistral-7B-v0.3 \citep{jiang2023mistral7b}) and tasks (multiple-choice and open-ended generation), significantly reducing attribute conflicts and achieving superior performance across multiple attributes datasets (e.g., concurrent improvements on TruthfulQA(+13\%), BBQ(+4\%)).  Additionally, MSRS generalizes well to standard NLP tasks, achieving gains on HellaSwag (+3.8\%) and GLUE (+4.9\%). Our contributions can be summarized as follows:
\vspace{-8pt}
\begin{itemize}
    \item We develop MSRS, a novel multi-subspace representation fine-tuning method that mitigates interference between distinct attribute steering within task-specific subspaces while capturing shared attribute directions in a common subspace. This design facilitates effective integration of multiple attribute steering objectives, enabling synergistic control over LLM behavior.
    \item MSRS demonstrates effective scalability across multiple attributes, maintaining performance gains in practical multi-attribute scenarios.
    \item Our method demonstrates significant performance improvements over other steering approaches on tasks such as multiple-choice and open-ended generation.
\end{itemize}

\vspace{-8pt}
\section{Related Work}
\textbf{Activation Steering Methods.} Activation steering guides model behavior by adjusting internal activations without modifying parameters~\citep{im2025unified}. Early approaches like CAA~\citep{rimsky-etal-2024-steering} and ITI~\citep{NEURIPS2023_81b83900} rely on static vectors derived from contrastive pairs or specific attention heads. While recent works like ACT~\citep{wang2025adaptive} and MAT-Steer~\citep{nguyen2025multi} incorporate clustering or orthogonal constraints to handle multiple patterns, they primarily focus on individual attributes or simple vector combinations. However, previous methods primarily address steering for individual attributes or rely on simple combinations of steering vectors. To solve these limitations, we focus on mitigating interference and optimizing composability across multiple attributes.

\noindent \textbf{Representation Fine-Tuning Methods.} Unlike single-vector steering, representation fine-tuning learns higher-rank transformations to enhance control expressivity~\citep{wu2024reft}. Methods such as LoFIT~\citep{yin2024lofitlocalizedfinetuningllm} offer targeted adjustments by modifying critical attention heads. More recently, CS-ReFT~\citep{zhou2025compositionalsubspacerepresentationfinetuning} advanced this by learning orthonormal subspaces for distinct skills composed via a router. However, while CS-ReFT emphasizes task isolation via routing, our approach focuses on the adaptive integration of attribute-specific subspaces. Unlike previous methods that train steering functions in the same space, we aim to develop representation fine-tuning methods to tune different attribute-specific subspaces and achieve the adaptive integration of multiple attribute steering spaces. See more details in Appendix \ref{app:related_work}.

\section{Motivation}
To better illustrate our motivation and approach, we first revise  ReFT~\citep{wu2024reft}. In ReFT, it aims to steer the hidden representation $h \in \mathbb{R}^d$ by fine-tuning an $r$-dimensional subspace spanned by the rows of $R$. Specifically, we can define the intervention function  $\Phi$ as:
\begin{equation}
\Phi(h; R, W, b) = h + R^{\top} \left(W h + b - R h\right), 
\label{eq:steering_func}
\end{equation}
where the learned low-rank projection matrix $R \in \mathbb{R}^{r \times d}$, is typically constrained to have orthonormal rows ($R R^{\top} = I_r$), and $W \in \mathbb{R}^{r \times d}$, $b \in \mathbb{R}^r$ are trainable parameters. $\Phi(h; R, W, b)$ is integrated into the model's representations to guide the output towards desired attributes. The steering-affected output is subsequently optimized to minimize the target objective, thereby refining the parameters $W$, $b$, and $R$. This method enables efficient steering of model representations by manipulating the hidden activations in a learned, low-rank subspace. 

While ReFT encounters significant limitations when applied to multi-attribute scenarios. ReFT assumes a single attribute per input, but real-world inputs often involve multiple attributes. Training the matrix \( R \) on such multi-attribute inputs forces all steering directions into the same space, causing interference that hinders the model's ability to balance the needs of each attribute, ultimately limiting its performance across all attributes. ~\citet{zhou2025compositionalsubspacerepresentationfinetuning} attempts to address this by partitioning \( R \) into equal-sized subspaces, each dedicated to a specific attribute, in an effort to reduce interference. However, this approach overlooks the fact that different attributes require subspaces of varying sizes based on their expressive needs. As a result, attributes with higher complexity may not receive enough capacity for effective steering, while simpler attributes may waste valuable space. 

To address these challenges, we propose Multi-Subspace Representation Steering (MSRS). MSRS mitigates the interference between attributes by assigning attribute-specific orthogonal subspaces and adapts the size of each subspace to fit the expressive needs of the corresponding attribute by utilizing Singular Value Decomposition (SVD), enabling dynamic adjustment for efficient representational space use. Furthermore, we introduce a shared subspace that captures common steering directions across attributes while learning the intricate interactions between them. This shared subspace enables the model to learn complex combinatory relationships between attributes, offering a more flexible and effective integration than methods that simply use gating mechanisms to combine attribute-specific subspaces.

\section{Methodology}

\subsection{Multi-Attribute Steering Direction Extraction}
\label{sec:multi-attribute}
To enable precise and simultaneous control over multiple attributes, we extract steering directions that disentangle shared and attribute-specific features, identifying significant directions in the activation space.


\noindent \textbf{Attribute-wise Activation Aggregation.} To extract the primary steering directions for each attribute from the activation values, we first capture the key feature representations for each attribute. Specifically, for each attribute \( i \), we compute the average activation \( \tau_i \) from its corresponding dataset \( \mathcal{D}_i \). We extract the model's intermediate activation \( h_{i,j}^l \) of layer \( l \) for each sample \( j \) at the last few tokens, specifically averaging the activations from these tokens, as they capture the full context of the prompt ~\citep{lei2025representation}. The average activation for attribute \( i \) is then computed as \( \tau_i = \frac{1}{|\mathcal{D}_i|} \sum_{j=1}^{|\mathcal{D}_i|} h_{i,j}^l \). To integrate information across all \( n \) attributes, we construct a combined activation matrix \( \tau_c = [\tau_1 \; | \; \tau_2 \; | \; \dots \; | \; \tau_n] \in \mathbb{R}^{d \times n} \).

\noindent \textbf{Shared and Specific Subspace Extraction.} To retain common knowledge while enabling attribute-specific steering, we perform singular value decomposition (SVD) on the aggregated activation matrix \( \tau_c \), i.e., \( \tau_c = U_c \Sigma_c V_c^\top \). We adaptively select the smallest number \( r_s \) such that the cumulative energy (sum of top \( r_s \) singular values) accounts for at least 60\% of the total energy in \( \Sigma_c \). This yields the shared subspace of \( U_c \), defined as \( B_{\text{shared}} = U_{c, 1:r_s}^\top \in \mathbb{R}^{r_s \times d} \). Intuitively, \( B_{\text{shared}} \) captures the dominant shared directions across all attributes. 

For each attribute \( i \), we then iteratively isolate attribute-specific directions by projecting \( H_i^{(i)} \) onto both the shared subspace and all previously extracted private subspaces, computing the residual, i.e.,
\begin{equation}
    H_{\text{res}}^{(i)} = H_i^{(i)} - B_{\text{shared}}^\top B_{\text{shared}} H_i^{(i)} - \sum_{k=1}^{i-1} B_k^\top B_k H_i^{(i)},
\end{equation}
where \( H_i^{(i)} \) denotes the activation matrix for attribute \( i \), formed by concatenating the sample representations, i.e., \( H_i^{(i)} = \left[ h_{i,1}^l, h_{i,2}^l, \dots, h_{i,D_i}^l \right] \). Applying SVD to \( H_{\text{res}}^{(i)} \), we obtain \( H_{\text{res}}^{(i)} = U^{(i)} S^{(i)} V^{(i)\top} \). Similarly, we select the smallest number \( r_i \) such that the top \( r_i \) singular values of \( S^{(i)} \) capture the principal energy, and define the private subspace as \( B_i = \left(U^{(i)}_{1:r_i} \right)^\top \in \mathbb{R}^{r_i \times d} \). Generally, \( B_i \) captures directions orthogonal to the shared subspace and prior attributes, preserving strictly unique attribute-specific semantics. By selecting the top singular vectors, we capture high-variance directions that represent the most expressive steering dimensions. It allows us to automatically allocate varying subspace sizes for each attribute based on its expressive needs. For complex attributes, it selects more top vectors, while simpler attributes are allocated smaller subspaces, effectively addressing the mismatch between each attribute's steering capacity and its allocated subspace size.
This adaptive allocation allows each attribute subspace to retain only as much representational capacity as needed, reflecting its inherent complexity.

The alignment matrix \( S_{\text{align}} \) is constructed by concatenating the shared and private subspace bases: \begin{equation} S_{\text{align}} = [B_{\text{shared}}, B_1, B_2, \dots, B_n] \in \mathbb{R}^{(r_s + \sum_{i=1}^n r_i) \times d}. \label{eq:salign} \end{equation}

\subsection{Adaptive Subspace Selecting}
\label{sec:adaptive-choosing}
\vspace{-2pt}To steer multiple attributes effectively, it is crucial to avoid the interference that arises when steering vectors for different attributes are trained in the same space. Furthermore, traditional methods often rely on summing or averaging these vectors, which often fail to produce an effective combination, as different attributes may require different subspace sizes or levels of emphasis. In contrast, we propose an adaptive mechanism that enables the model to train in a specific subspace and learn to combine different steering subspaces optimally, overcoming prior limitations. 

Based on~\eqref{eq:steering_func}, we introduce a mask network \( m(h) = \text{sigmoid}(\text{MLP}(h)) \in [0, 1]^r \), which assigns weights to each subspace dimension. The intervention function becomes:
\begin{equation}
\Phi_{l,p}(h; R, W, b,m) = h + R^\top  \text{diag}(m(h))  (W h + b - R h), 
\end{equation}
where \( \text{diag}(m(h)) \in \mathbb{R}^{r \times r} \) is a diagonal matrix.

\subsection{Optimization Objective} 
\vspace{-4pt}We optimize the steering function \( \Phi_{l,p}(h; R, W, b, m) \), applying it to the representation \( H_{l,p} \) at layer \( l \) and position \( p \). This changes the representation and influences the model's output, which is then used to compute the task-specific loss \( \mathcal{L}_{\text{task}} \), defined as the standard cross-entropy loss between the predicted logits and the ground truth labels, reflecting the model's performance on the downstream task: \( \mathcal{L}_{\text{task}} = \text{CrossEntropy}\left( \text{Softmax}\left( \Phi_{l,p}(h) \right), y \right) \).

To enable the steering function to perform meaningful and disentangled attribute control, we introduce a subspace regularization term. Specifically, to encourage adaptive selection of relevant subspaces, we define a binary prior mask \( m_{\text{prior}} \in \{0, 1\}^r \), where entries corresponding to the shared subspace \( B_{\text{shared}} \) and the attribute-specific subspace \( B_i \) are set to 1, and all others to 0. The regularization loss is defined as \( \mathcal{L}_{\text{reg}} = \left\| m(h) - m_{\text{prior}} \right\|_2^2 \), which encourages the model to steer primarily within subspaces that are relevant to the target attribute, while suppressing activation in unrelated dimensions.

We further encourage the learned representation $ R $, which is initialized to match the dimensions of $ S_{\text{align}} $, to align with the structured basis $ S_{\text{align}} $, as defined in~\eqref{eq:salign}.
The alignment loss is defined as
\(\mathcal{L}_{\text{align}} = 1 - \frac{\langle R, S_{\text{align}} \rangle_F}{\| R \|_F \| S_{\text{align}} \|_F}\),
where \( \langle A, B \rangle_F = \text{Tr}(A^\top B) = \sum_{i,j} A_{ij} B_{ij} \) denotes the Frobenius inner product, and \( \| \cdot \|_F \) represents the Frobenius norm. This formulation encourages \( R \) to lie in the subspace spanned by both shared and attribute-specific directions, promoting more controllable and semantically meaningful representations during training. 

The overall optimization objective is as follows: 
\begin{equation}
\mathcal{L} = \mathcal{L}_{\text{task}} + \lambda_1 \mathcal{L}_{\text{reg}} + \lambda_2 \mathcal{L}_{\text{align}},
\end{equation}
where \( \lambda_1, \lambda_2 > 0 \) are hyperparameters balancing the terms. This optimization guarantees both attribute-wise subspace alignment and inter-attribute separation, ultimately yielding an effective steering space capable of precise and disentangled multi-attribute control. By integrating different subspaces with a weighting network, we enable adaptive subspace combination, alleviating trade-offs and optimizing performance across diverse attributes' steering. 

\begin{table*}[t]
\centering
\vspace{-0pt}
\setlength{\tabcolsep}{1mm}
\caption{Evaluation results on TruthfulQA, BBQ, Alpaca, Refusal, and HelpSteer.}
\resizebox{\textwidth}{!}{%
\begin{tabular}{lllllllllllll}
\hline\hline
\textbf{Method} & \multicolumn{4}{c}{\textbf{TruthfulQA}} & \textbf{BBQ} & \textbf{Alpaca} & \textbf{Refusal} & \multicolumn{3}{c}{\textbf{HelpSteer}} \\
& MC1~($\uparrow$) & MC2~($\uparrow$) & Bleu~($\uparrow$) & Bleurt~($\uparrow$) & Acc~($\uparrow$) & Win~($\uparrow$) & Sorry~($\uparrow$) & Help.~($\uparrow$) & Coh.~($\uparrow$) & Ver.~($\downarrow$)
\\
\hline
Llama3-8b-inst. & 27.47$_{\pm 0.21}$ & 45.63$_{\pm 0.28}$ & 46.89$_{\pm 0.25}$ & 57.88$_{\pm 0.33}$ & 0.608$_{\pm 0.011}$ & 0.12$_{\pm 0.010}$ & 0.491$_{\pm 0.021}$ & 3.76$_{\pm 0.03}$ & 3.41$_{\pm 0.04}$ & 2.33$_{\pm 0.03}$ \\
ICL & 28.37$_{\pm 0.25}$ & 46.21$_{\pm 0.26}$ & 48.35$_{\pm 0.23}$ & 60.44$_{\pm 0.31}$ & 0.619$_{\pm 0.013}$ & 0.27$_{\pm 0.011}$ & \underline{0.521}$_{\pm 0.024}$ & 3.82$_{\pm 0.02}$ & 3.64$_{\pm 0.03}$ & 2.41$_{\pm 0.04}$ \\
CAA & 28.41$_{\pm 0.18}$ & 47.55$_{\pm 0.23}$ & 45.42$_{\pm 0.21}$ & 61.54$_{\pm 0.28}$ & 0.629$_{\pm 0.009}$ & 0.29$_{\pm 0.008}$ & 0.493$_{\pm 0.018}$ & 3.77$_{\pm 0.02}$ & \underline{3.89}$_{\pm 0.02}$ & 2.51$_{\pm 0.03}$ \\
ITI & \textbf{36.50}$_{\pm 0.20}$ & \underline{54.29}$_{\pm 0.29}$ & 43.22$_{\pm 0.24}$ & \underline{66.30}$_{\pm 0.30}$ & 0.612$_{\pm 0.012}$ & 0.23$_{\pm 0.009}$ & 0.280$_{\pm 0.015}$ & 3.82$_{\pm 0.02}$ & 3.21$_{\pm 0.04}$ & 2.06$_{\pm 0.03}$ \\
ReFT & 29.58$_{\pm 0.16}$ & 49.51$_{\pm 0.22}$ & 52.08$_{\pm 0.18}$ & 64.06$_{\pm 0.29}$ & 0.637$_{\pm 0.008}$ & \underline{0.30}$_{\pm 0.008}$ & 0.451$_{\pm 0.020}$ & 3.78$_{\pm 0.02}$ & 3.83$_{\pm 0.02}$ & 2.38$_{\pm 0.02}$ \\
CS-ReFT & 34.82$_{\pm 0.17}$ & 53.23$_{\pm 0.24}$ & \underline{52.18}$_{\pm 0.19}$ & 63.26$_{\pm 0.28}$ & \underline{0.641}$_{\pm 0.008}$ & 0.21$_{\pm 0.007}$ & 0.514$_{\pm 0.022}$ & 3.78$_{\pm 0.04}$ & 3.86$_{\pm 0.03}$ & \textbf{2.01}$_{\pm 0.02}$ \\
MAT-STEER & 29.29$_{\pm 0.18}$ & 49.67$_{\pm 0.25}$ & 43.81$_{\pm 0.22}$ & 55.31$_{\pm 0.34}$ & 0.622$_{\pm 0.010}$ & 0.14$_{\pm 0.009}$ & 0.420$_{\pm 0.019}$ & \underline{3.84}$_{\pm 0.02}$ & 3.63$_{\pm 0.03}$ & 2.29$_{\pm 0.03}$ \\
\rowcolor{blue!10} \textbf{MSRS} & \underline{34.91}$_{\pm 0.12}$ & \textbf{56.32}$_{\pm 0.20}$ & \textbf{52.32}$_{\pm 0.14}$ & \textbf{66.75}$_{\pm 0.26}$ & \textbf{0.645}$_{\pm 0.010}$ & \textbf{0.36}$_{\pm 0.008}$ & \textbf{0.529}$_{\pm 0.02}$ & \textbf{3.89}$_{\pm 0.01}$ & \textbf{3.96}$_{\pm 0.02}$ & \underline{2.04}$_{\pm 0.02}$ \\
\hline
Qwen2-7b-inst. & 26.38$_{\pm 0.20}$ & 45.41$_{\pm 0.29}$ & 49.63$_{\pm 0.24}$ & 65.28$_{\pm 0.32}$ & 0.634$_{\pm 0.009}$ & 0.12$_{\pm 0.009}$ & 0.384$_{\pm 0.018}$ & 3.51$_{\pm 0.03}$ & 3.80$_{\pm 0.02}$ & 2.28$_{\pm 0.03}$ \\
ICL & 26.84$_{\pm 0.24}$ & 48.33$_{\pm 0.27}$ & 49.79$_{\pm 0.22}$ & 66.63$_{\pm 0.30}$ & 0.633$_{\pm 0.010}$ & 0.16$_{\pm 0.010}$ & 0.413$_{\pm 0.020}$ & 3.65$_{\pm 0.03}$ & \textbf{3.88}$_{\pm 0.02}$ & 2.40$_{\pm 0.03}$ \\
CAA & 28.44$_{\pm 0.17}$ & 47.25$_{\pm 0.25}$ & 48.35$_{\pm 0.20}$ & 58.97$_{\pm 0.35}$ & 0.635$_{\pm 0.008}$ & 0.26$_{\pm 0.008}$ & 0.404$_{\pm 0.019}$ & \underline{3.73}$_{\pm 0.02}$ & \underline{3.87}$_{\pm 0.02}$ & \underline{2.20}$_{\pm 0.02}$ \\
ReFT & \underline{29.83}$_{\pm 0.15}$ & 48.69$_{\pm 0.24}$ & 52.57$_{\pm 0.17}$ & 71.15$_{\pm 0.26}$ & 0.636$_{\pm 0.008}$ & \underline{0.43}$_{\pm 0.007}$ & 0.421$_{\pm 0.017}$ & 3.63$_{\pm 0.03}$ & 3.78$_{\pm 0.02}$ & 2.38$_{\pm 0.03}$ \\
CS-ReFT & 29.15$_{\pm 0.18}$ & 48.92$_{\pm 0.25}$ & \underline{52.85}$_{\pm 0.20}$ & 72.25$_{\pm 0.28}$ & 0.639$_{\pm 0.008}$ & 0.41$_{\pm 0.009}$ & 0.425$_{\pm 0.018}$ & 3.72$_{\pm 0.03}$ & 3.84$_{\pm 0.02}$ & 2.22$_{\pm 0.02}$ \\
MAT-STEER & 23.08$_{\pm 0.19}$ & \underline{49.51}$_{\pm 0.28}$ & 51.72$_{\pm 0.19}$ & \underline{72.53}$_{\pm 0.25}$ & \underline{0.641}$_{\pm 0.007}$ & 0.18$_{\pm 0.009}$ & \underline{0.429}$_{\pm 0.016}$ & 3.70$_{\pm 0.02}$ & 3.77$_{\pm 0.02}$ & 2.25$_{\pm 0.02}$ \\
\rowcolor{blue!10} \textbf{MSRS} & \textbf{34.72}$_{\pm 0.11}$ & \textbf{53.27}$_{\pm 0.21}$ & \textbf{53.10}$_{\pm 0.12}$ & \textbf{74.90}$_{\pm 0.21}$ & \textbf{0.642}$_{\pm 0.006}$ & \textbf{0.45}$_{\pm 0.007}$ & \textbf{0.445}$_{\pm 0.011}$ & \textbf{3.76}$_{\pm 0.01}$ & 3.82$_{\pm 0.01}$ & \textbf{2.17}$_{\pm 0.02}$ \\
\hline
Mistral-7b-v0.3 & 18.83$_{\pm 0.28}$ & 36.54$_{\pm 0.33}$ & 41.56$_{\pm 0.29}$ & 54.52$_{\pm 0.38}$ & 0.614$_{\pm 0.014}$ & 0.14$_{\pm 0.012}$ & 0.631$_{\pm 0.028}$ & 3.75$_{\pm 0.03}$ & 3.92$_{\pm 0.02}$ & 2.36$_{\pm 0.04}$ \\
ICL & 21.92$_{\pm 0.26}$ & 49.23$_{\pm 0.30}$ & 44.11$_{\pm 0.27}$ & 57.64$_{\pm 0.35}$ & 0.622$_{\pm 0.013}$ & 0.18$_{\pm 0.011}$ & 0.642$_{\pm 0.025}$ & 3.77$_{\pm 0.03}$ & \textbf{3.94}$_{\pm 0.01}$ & 2.39$_{\pm 0.03}$ \\
CAA & 28.77$_{\pm 0.19}$ & \underline{52.05}$_{\pm 0.26}$ & \textbf{53.85}$_{\pm 0.20}$ & 62.27$_{\pm 0.32}$ & \textbf{0.646}$_{\pm 0.009}$ & 0.22$_{\pm 0.009}$ & 0.663$_{\pm 0.021}$ & 3.76$_{\pm 0.02}$ & 3.83$_{\pm 0.02}$ & \underline{2.27}$_{\pm 0.02}$ \\
ReFT & \underline{30.07}$_{\pm 0.17}$ & 49.69$_{\pm 0.28}$ & 49.39$_{\pm 0.23}$ & 66.01$_{\pm 0.31}$ & 0.614$_{\pm 0.011}$ & \underline{0.32}$_{\pm 0.008}$ & \underline{0.668}$_{\pm 0.019}$ & \underline{3.80}$_{\pm 0.02}$ & 3.85$_{\pm 0.02}$ & 2.33$_{\pm 0.03}$ \\
CS-ReFT & 29.45$_{\pm 0.20}$ & 51.12$_{\pm 0.27}$ & 49.95$_{\pm 0.22}$ & \underline{69.20}$_{\pm 0.33}$ & 0.638$_{\pm 0.010}$ & 0.31$_{\pm 0.010}$ & 0.659$_{\pm 0.020}$ & 3.79$_{\pm 0.02}$ & 3.89$_{\pm 0.02}$ & 2.29$_{\pm 0.02}$ \\
MAT-STEER & 25.45$_{\pm 0.22}$ & 48.38$_{\pm 0.31}$ & 49.46$_{\pm 0.24}$ & 62.62$_{\pm 0.34}$ & 0.631$_{\pm 0.010}$ & 0.19$_{\pm 0.010}$ & 0.644$_{\pm 0.023}$ & 3.78$_{\pm 0.02}$ & 3.86$_{\pm 0.02}$ & 2.29$_{\pm 0.02}$ \\
\rowcolor{blue!10} \textbf{MSRS} & \textbf{31.32}$_{\pm 0.11}$ & \textbf{52.62}$_{\pm 0.13}$ & \underline{50.61}$_{\pm 0.08}$ & \textbf{71.39}$_{\pm 0.17}$ & \underline{0.644}$_{\pm 0.007}$ & \textbf{0.38}$_{\pm 0.006}$ & \textbf{0.693}$_{\pm 0.013}$ & \textbf{3.82}$_{\pm 0.02}$ & \underline{3.93}$_{\pm 0.01}$ & \textbf{2.21}$_{\pm 0.02}$ \\
\hline\hline
\end{tabular}}
\label{tab:main-results}
\vspace{-12pt}
\end{table*}

\section{Experiments} 
\subsection{Experimental Settings}
\noindent {\bf Datasets and Metrics.} We evaluate MSRS on three dataset pairings designed to assess multi-attribute steering trade-offs (details in Appendix~\ref{appendix:algorithm-a2}):
\textbf{1) TruthfulQA \& BBQ}: We measure \emph{truthfulness} via TruthfulQA metrics (MC1/2, BLEU, BLEURT)~\citep{lin-etal-2022-truthfulqa} and \emph{social bias} via BBQ accuracy~\citep{parrish-etal-2022-bbq}.
\textbf{2) Alpaca \& Refusal}: We evaluate \emph{instruction following} using Alpaca win-rates (vs. text-davinci-003)~\citep{alpaca,alpaca_eval} and \emph{safety refusal} via Sorry-Bench scores (judged by Mistral-7B-v0.2)~\citep{xie2025sorrybench}.
\textbf{3) HelpSteer}: We assess \emph{helpfulness}, \emph{coherence}, and \emph{verbosity} on a 0–4 scale using a GPT-4-o judge~\citep{nguyen2025multi,wang2023helpsteer}.
To ensure steering does not degrade general capabilities, we also report \textbf{General Utility} accuracy across standard benchmarks: Hellaswag~\citep{zellers-etal-2019-hellaswag}, RACE~\citep{lai-etal-2017-race}, MMLU~\citep{hendrycks2020measuring}, OpenBookQA~\citep{mihaylov-etal-2018-suit}, and GLUE~\citep{wang2018glue}.

\vspace{-2pt}\noindent {\bf Models and Baselines.}
We evaluate MSRS on 4 models: Llama2-7B \citep{touvron2023llama}, Llama3-8B-Instruct \citep{grattafiori2024llama}, Qwen2-7B-Instruct \citep{team2024qwen2} , Mistral-7B-v0.3 \citep{jiang2023mistral7b}.
And we compare MSRS with 6 baselines, grouped into 3 categories: \textbf{1) In-context Learning} \citep{brown2020language}: Utilizes prompts to steer attributes without altering model parameters. \textbf{2) Fine-tuning Methods}:
\textbf{ReFT} \citep{wu2024reft}, which adjusts model representations by fine-tuning the representation to align with target attributes.
\textbf{CS-ReFT} \citep{zhou2025compositionalsubspacerepresentationfinetuning}, which employs a lightweight router to compose task-specific orthonormal subspaces for interference-free adaptation.
\textbf{3) Steering Methods}: \textbf{ITI} \citep{NEURIPS2023_81b83900} applies inference-time interventions to modify activations and guide model outputs. \textbf{CAA} \citep{rimsky-etal-2024-steering} steers behavior by injecting contrastive activation vectors derived from positive and negative examples.
\textbf{MAT-STEER} \citep{nguyen2025multi} implements multi-attribute steering with orthogonal constraints to minimize interference between attributes.

\noindent \textbf{Experimental Setup.} All experiments were conducted on NVIDIA V100 GPUs. We employed the Adam optimizer with a learning rate of \(9 \times 10^{-4}\) and a batch size of 4. The total subspace rank \( R \) is set to match the rank of \( S_{\text{align}} \) , with dual regularization coefficients \(\lambda_1 =0.3\), \(\lambda_2 = 0.5\) (robustness verified in Appendix \ref{sec:hyperparam-sensitivity}). Following the ReFT setting, we fixed the intervention to the last 5 tokens. For each configuration, we report the average and standard deviation over 3 runs with random seeds \{42, 43, 44\}. 

\subsection{Main results}
\vspace{-2pt}\textbf{MSRS excels in multiple-choice tasks.} We first evaluate MSRS on TruthfulQA and BBQ to target the trade-off between truthfulness and bias. 
As shown in Table~\ref{tab:main-results}, baseline methods often fail to optimize both attributes simultaneously. For example, ITI improves truthfulness (MC1 of 36.50 on Llama3-8B-Instruct) but sacrifices bias mitigation (BBQ Acc of 0.612), while CAA boosts bias (BBQ Acc of 0.646 on Mistral-7B) at the cost of truthfulness (MC1 of 28.77). ICL shows moderate improvements across metrics but lacks standout performance, and MAT-STEER enhances MC and Acc, but its BLEU and BLEURT scores drop. Unlike these methods, MSRS consistently balances both attributes, achieving superior performance across both attributes on multiple models. These results demonstrate MSRS’s ability to jointly optimize conflicting objectives across all models.

\noindent \textbf{MSRS demonstrates strong performance in open-ended generation tasks.} We assess MSRS on Alpaca, Refusal, and HelpSteer datasets to evaluate trade-offs between instruction following, refusal, and output quality attributes. 
As shown in Table~\ref{tab:main-results}, MSRS demonstrates superior capability in balancing instruction-following and refusal. On Llama3-8B-Instruct, it effectively surpasses strong baselines like ReFT on Alpaca and CAA on Sorry-Bench, whereas methods like ITI suffer from significant degradation in refusal performance. Regarding HelpSteer, MSRS consistently enhances Helpfulness, Coherence, and Verbosity. In contrast, other baselines  exhibit clear trade-offs, prioritizing specific metrics (e.g., Verbosity) at the expense of others. Overall, MSRS achieves comprehensive superiority; even in rare cases where it does not rank first, it maintains comparable performance with a negligible gap, avoiding the severe performance drops seen in other methods.

\begin{figure}[t]
    \centering
    \includegraphics[width=0.9\linewidth]{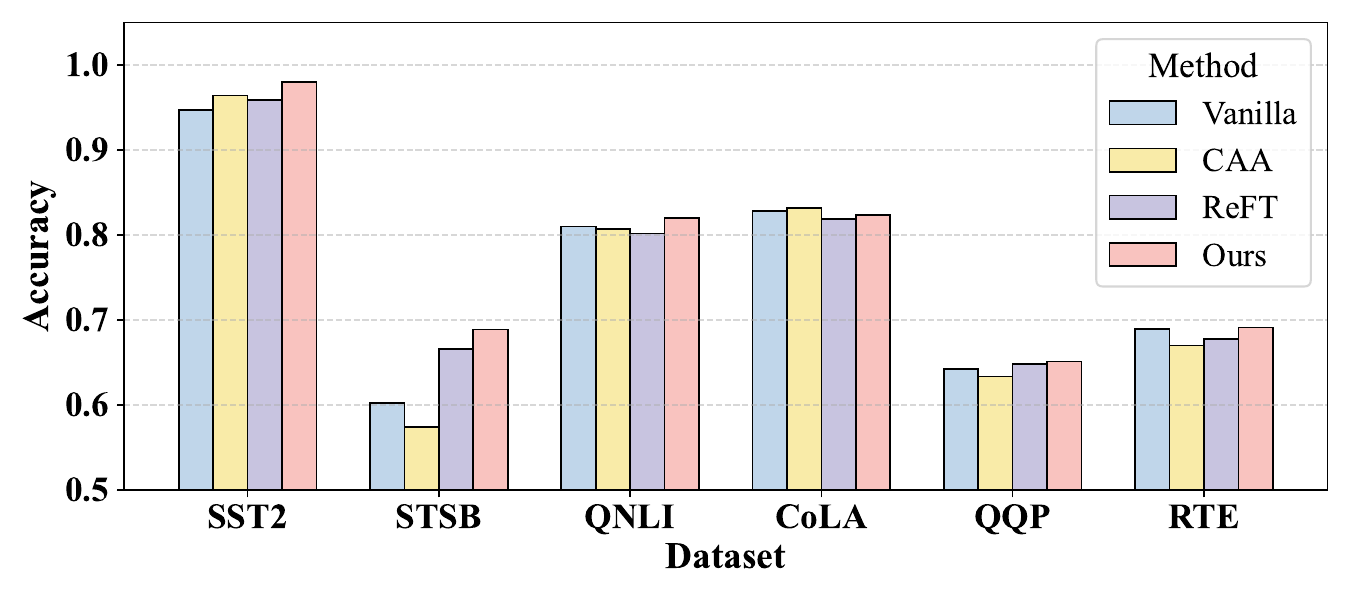}
    \caption{
    Comparison of model performance across GLUE.}
    \label{fig:glue-performance}
\vspace{-6pt}
\end{figure}

\begin{table}[t] 
\centering
\caption{General capabilities on several benchmarks.}
\label{tab:general-capability}

\renewcommand{\arraystretch}{1.0}

\resizebox{\linewidth}{!}{
    \begin{tabular}{lccccc}
    \hline\hline
    \textbf{Model / Method} & \textbf{HellaSwag} & \textbf{RACE} & \textbf{MMLU} & \textbf{OBQA} & \textbf{GLUE} \\
    \hline
    \multicolumn{6}{l}{\textit{Llama-3-8B-Instruct}} \\
    \hspace{3mm} Vanilla & 0.801 & 0.671 & 0.655 & 0.556 & 0.726 \\
    \rowcolor{blue!10} \hspace{3mm} + MSRS (Ours) & \textbf{0.839} & \textbf{0.683} & \textbf{0.657} & \textbf{0.568} & \textbf{0.775} \\
    \hline
    \multicolumn{6}{l}{\textit{Qwen2-7B-Instruct}} \\
    \hspace{3mm} Vanilla & 0.831 & 0.625 & 0.695 & 0.606 & 0.825 \\
    \rowcolor{blue!10} \hspace{3mm} + MSRS (Ours) & \textbf{0.835} & \textbf{0.648} & \textbf{0.702} & \textbf{0.616} & \textbf{0.832} \\
    \hline
    \multicolumn{6}{l}{\textit{Mistral-7B-v0.3}} \\
    \hspace{3mm} Vanilla & 0.862 & 0.678 & \textbf{0.618} & 0.602 & 0.681 \\
    \rowcolor{blue!10} \hspace{3mm} + MSRS (Ours) & \textbf{0.874} & \textbf{0.681} & 0.613 & \textbf{0.622} & \textbf{0.707} \\
    \hline\hline
    \end{tabular}
}\vspace{-6pt}
\end{table}

\noindent \textbf{MSRS maintains strong general capabilities on standard NLP benchmarks.} We evaluate MSRS on standard benchmarks (Table~\ref{tab:general-capability} and Appendix~\ref{appendix:algorithm-MMLU}) to assess its robustness. MSRS preserves the base model's utility, consistently achieving comparable or superior performance. Specifically, on the GLUE benchmark, MSRS attains an average score of 0.775, outperforming competing methods (see details in Appendix ~\ref{subsec:glue-performance}). Figure~\ref{fig:glue-performance} further highlights improvements in sentiment and semantic tasks. This superior generalization is driven by our shared subspace mechanism, which learns common steering directions applicable across tasks, thereby avoiding the negative trade-offs distinct in prior works.

\begin{figure}[t]
  \centering
  \vspace{-2pt}
  \includegraphics[width=\linewidth]{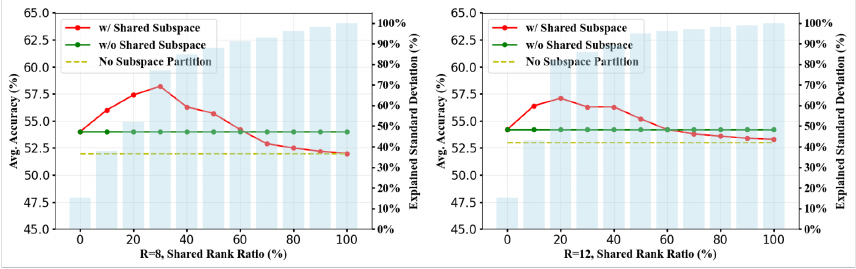}
  \caption{Relationship between performance and shared rank ratio, alongside the explained standard deviation.}
  \label{fig:ms-performance}
  \vspace{-10pt}
\end{figure}

\subsection{Core Multi-Subspace Design}

\textbf{The shared subspace facilitates the integration of multi-attribute features, resulting in enhanced overall performance.
} Since the effectiveness of MSRS depends on the shared subspace, we investigate the impact of the rank ($k$ of the shared subspace) on performance, by exploring the allocation of the shared subspace under the condition of the same total subspace rank $R$. As shown in Figure \ref{fig:ms-performance}, we report the relationship between the performance (Avg. Accuracy) of the attributes and the proportion of the shared subspace within the total space (shared rank ratio).

Models with a shared subspace in $R$ outperform those without shared subspace (w/o Shared Subspace, only attribute-specific space) or with no subspace partition. The optimal shared subspace ratio varies with $R$: for $R = 8$, it is around 30\%, and for $R = 12$, it is about 20\%. As the shared subspace ratio increases, the attribute-specific space decreases, causing performance to decline. At 100\%, the model degenerates to the no subspace partition case. Additionally, we display the explained standard deviation (blue bars, i.e., the ratio of preserved singular values $\sigma$ to the total sum of singular values $\Sigma$). The optimal shared subspace ratio is between 20\% and 40\%, where the explained standard deviation more than 60\%. Accordingly, we choose the fewest singular vectors that capture at least 60\% of the total energy in $\Sigma_c$. See sensitivity analysis in Appendix~\ref{appendix:shared-subspace-analysis}.

\begin{table}[t]
\caption{Comparison of steering subspace training strategies.}
\label{tab:ablation-llama3}
\centering
\renewcommand{\arraystretch}{1.1} 

\resizebox{\linewidth}{!}{
\begin{tabular}{lccccc}
\toprule
 & \textbf{TruthfulQA} & \textbf{BBQ} & \textbf{Alpaca} & \textbf{Refusal} & \textbf{HelpSteer} \\
\cmidrule(lr){2-2} \cmidrule(lr){3-3} \cmidrule(lr){4-4} \cmidrule(lr){5-5} \cmidrule(lr){6-6}
\textbf{Method} & \textbf{Avg.} & \textbf{Acc} & \textbf{Win} & \textbf{Sorry} & \textbf{Avg.} \\
\midrule
Same Space & 39.55 & 0.637 & 0.30 & 0.451 & 3.88 \\
\rowcolor{blue!10} MSRS$_{\text{Rank}}$ & \textbf{43.12} & \textbf{0.646} & 0.35 & 0.527 & 3.92 \\
\rowcolor{blue!10} MSRS$_{\text{Attribute}}$ & 42.54 & 0.637 & \textbf{0.36} & \textbf{0.529} & \textbf{3.93} \\
\bottomrule
\end{tabular}
}\vspace{-8pt}
\end{table}

\noindent \textbf{Mechanism Analysis.} We compare three strategies for training steering subspaces:
(1) \textbf{Same Space}: All attributes are trained within a single global subspace without isolation.
(2) \textbf{$\text{MSRS}_{\text{Attribute}}$} (Default): The basis matrix $R$ is partitioned into semantic blocks $R = [ B_{\text{shared}} \,|\, B_1 \,|\, \dots \,|\, B_n ]$, where $m(h)$ modulates each block to adaptively activate specific attribute subspaces.
(3) \textbf{$\text{MSRS}_{\text{Rank}}$}: The basis $R$ is treated as $r$ independent vectors, with $m(h) \in \mathbb{R}^r$ weighting each rank individually for fine-grained control.
We adopt $\text{MSRS}_{\text{Attribute}}$ as the default implementation, as it offers the optimal balance between interpretability and effective attribute-level steering.
As shown in Table~\ref{tab:ablation-llama3}, the \textit{Same Space} baseline underperforms due to interference from conflicting objectives within a shared optimization space. MSRS mitigates this through subspace decoupling and adaptive modulation $m(h)$. Specifically, $\text{MSRS}_{\text{Attribute}}$ uses coarse-grained partitioning with dedicated attribute blocks, achieving consistent efficiency and gains in instruction-following tasks (e.g., Alpaca, Refusal). In contrast, $\text{MSRS}_{\text{Rank}}$ provides fine-grained control by modulating individual basis directions, yielding superior performance in tasks requiring precise feature suppression, such as Truthfulness and BBQ. (See Appendix~\ref{appendix:C}, Table~\ref{tab:ablation-space} for full results).

\subsection{Enhanced Steering Mechanisms}

\begin{figure}[t]
  \centering
  \includegraphics[width=\linewidth]{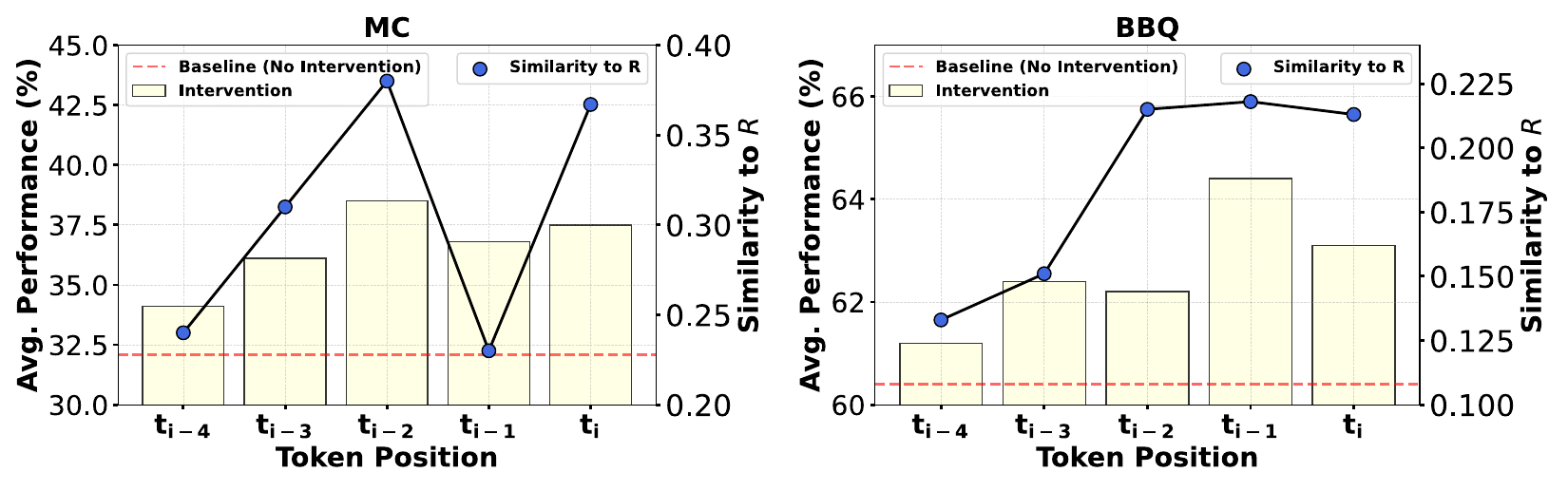}
  \caption{Token–$R$ similarity vs.\ performance under different token steering. Higher similarity correlates with better performance.}
  \label{fig:figure_layer}
  \vspace{-14pt} 
\end{figure}

\begin{figure}[t]
    \centering
    \includegraphics[width=0.85\linewidth]{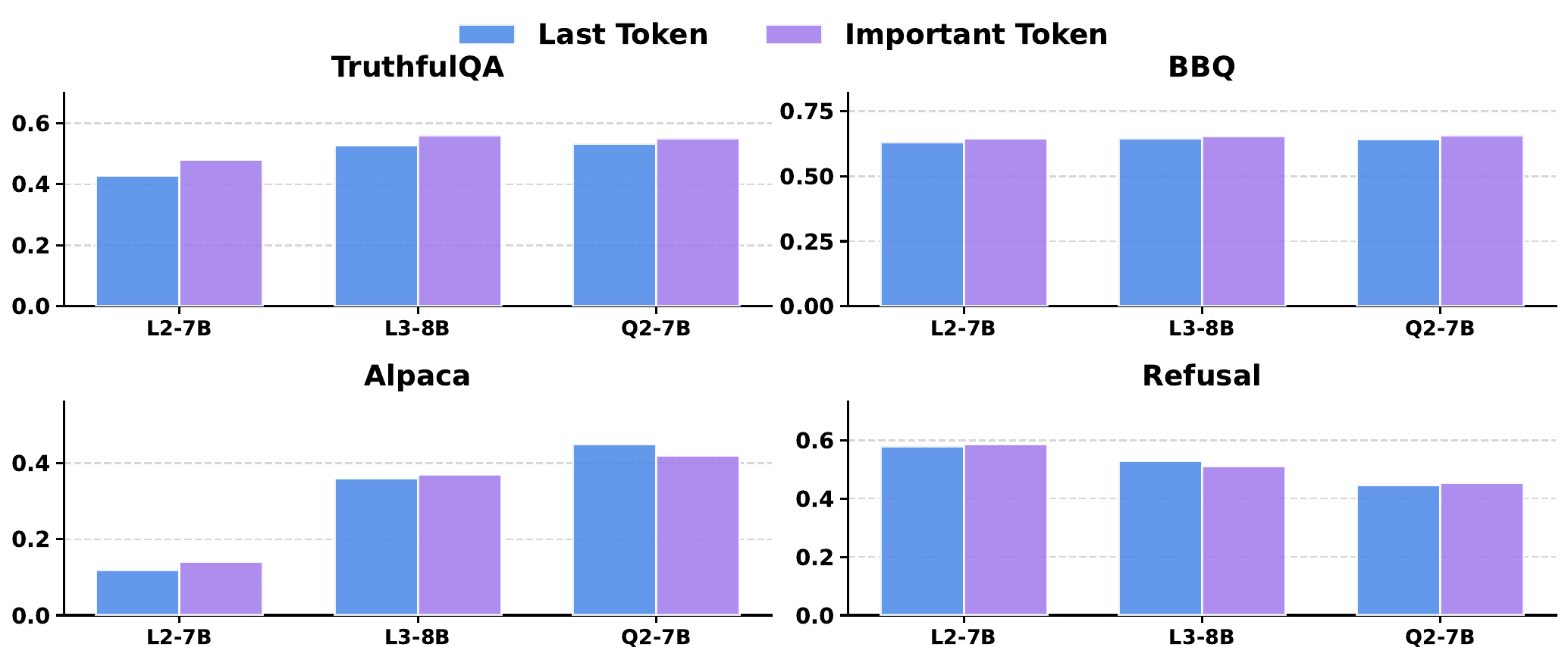}

    \vspace{-5pt} 
    
    \caption{Comparison of Last token vs. Important token steering. }
    \label{fig:ablation_study}
    \vspace{-18pt} 
\end{figure}

\begin{figure}[t]
    \centering
    \includegraphics[width=\linewidth]{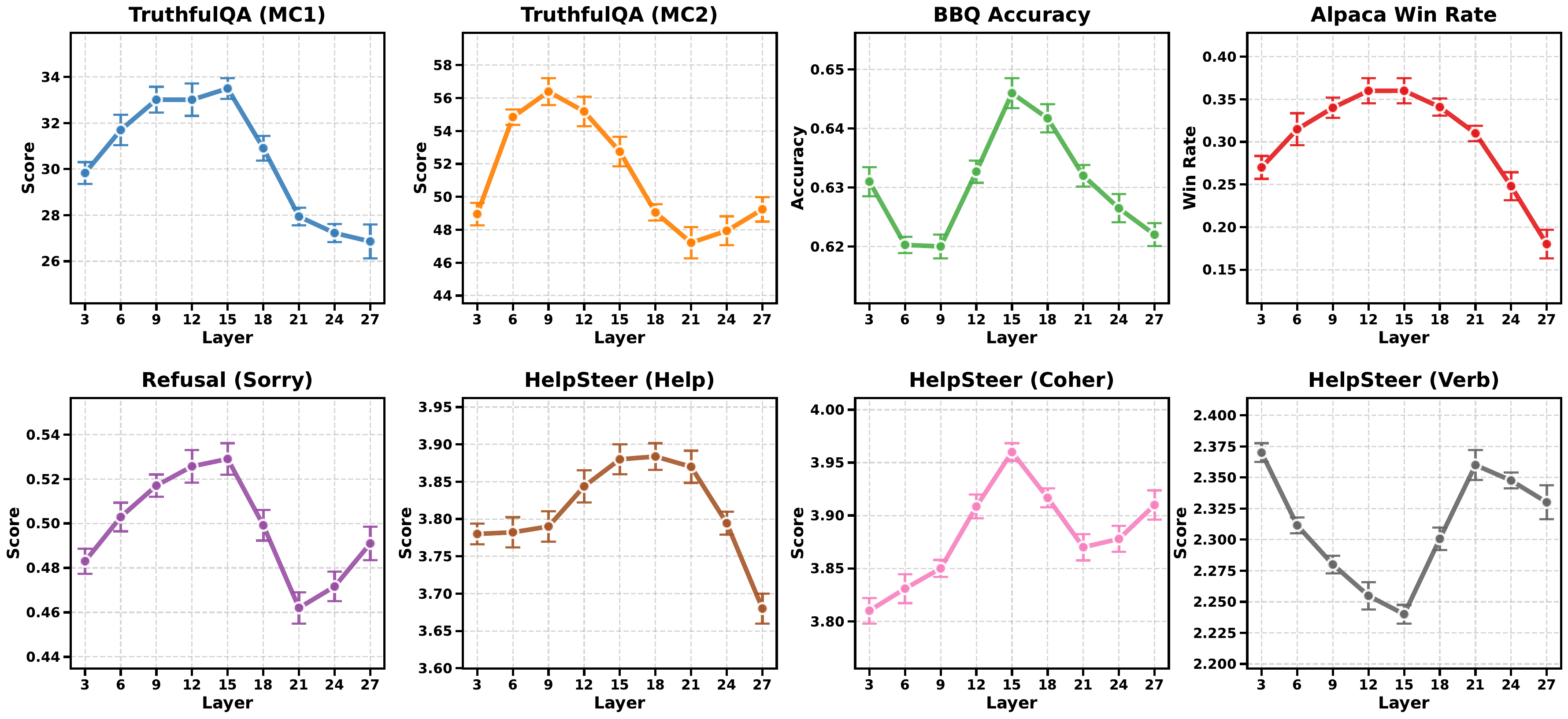} 
    \caption{
    Performance of interventions at different layers.}
    \label{fig:layer-sweep}
\vspace{-10pt}
\end{figure}

\begin{table}[t]
\centering
\caption{Inference-time comparison of different steering methods.}
\label{tab:inference-cost}
\renewcommand{\arraystretch}{1} 
\resizebox{\linewidth}{!}{
\begin{tabular}{lccc}
\toprule
\textbf{Method} & \textbf{Latency (ms/tok)} $\downarrow$ & \textbf{Throughput (tok/s)} $\uparrow$ & \textbf{VRAM (GB)} $\downarrow$ \\
\midrule
Baseline    & 102.44          & 9.76            & 16.10 \\
ReFT        & 115.72          & 8.64            & 16.26 \\
CAA         & 110.59          & 9.04            & 16.12 \\
\rowcolor{blue!10} MSRS (Ours) & 120.35 & 8.38  & 16.29 \\
\bottomrule
\end{tabular}
}\vspace{-8pt}
\end{table}

\noindent \textbf{Projection similarity serves as a proxy for steerability.} Our core insight is that steering efficacy varies across token positions; tokens aligning closely with the attribute-specific subspace ($R$) are semantically relevant, serving as optimal anchor points for intervention. To validate this, we examined the correlation between $R$-similarity and model performance by applying interventions at distinct single-token suffix positions (Figure \ref{fig:figure_layer}). The results empirically confirm our hypothesis: steering at tokens with higher similarity is more likely to yield superior performance. Moreover, the intervention outperforms the no-intervention baseline across all positions, demonstrating robust generalization. 

To quantify this capability, we conducted a comparative analysis between two intervention strategies: (1) \textit{Last Token}, where the steering vector is applied strictly to the final token of the sequence; and (2) \textit{Important Token}, which dynamically selects the intervention point from the recent context based on the highest projection similarity. As shown in Figure~\ref{fig:ablation_study}, the dynamic \textit{Important Token} strategy demonstrates competitive performance gains over the fixed \textit{Last Token} baseline. This indicates that the attribute-specific subspace derived by MSRS possesses the intrinsic capability to identify optimal anchor points for attribute intervention.


\textbf{Optimal Layer Selection.} We evaluate steering efficacy across different layers of  LLaMA3 (Figure~\ref{fig:layer-sweep}). Consistent with prior findings~\citep{skean2025layer}, mid-to-upper layers yield superior control, whereas lower layers lack semantic abstraction and deeper layers are prone to overfitting. Consequently, based on a grid search balancing attribute trade-offs on validation splits, we select layer 15 for LLaMA3, layer 9 for Qwen2, and layer 21 for Mistral for all subsequent experiments. Detailed results are provided in Appendix~\ref{appendix:E}.

\textbf{Computational Efficiency.} We evaluated the computational overhead of various methods using Llama3-8B-Instruct on the Alpaca dataset (sequence length 512, batch size 1). As shown in Table~\ref{tab:inference-cost}, MSRS introduces modest latency and VRAM overhead compared to the base model, with performance metrics remaining comparable to ReFT and only 
marginally higher than CAA. Appendix~\ref{app:complexity} further confirms the minimal overhead of the one-time offline initialization phase. Given that MSRS consistently delivers the superior
\begin{figure}[H]
    \centering
    \includegraphics[width=0.9\linewidth]{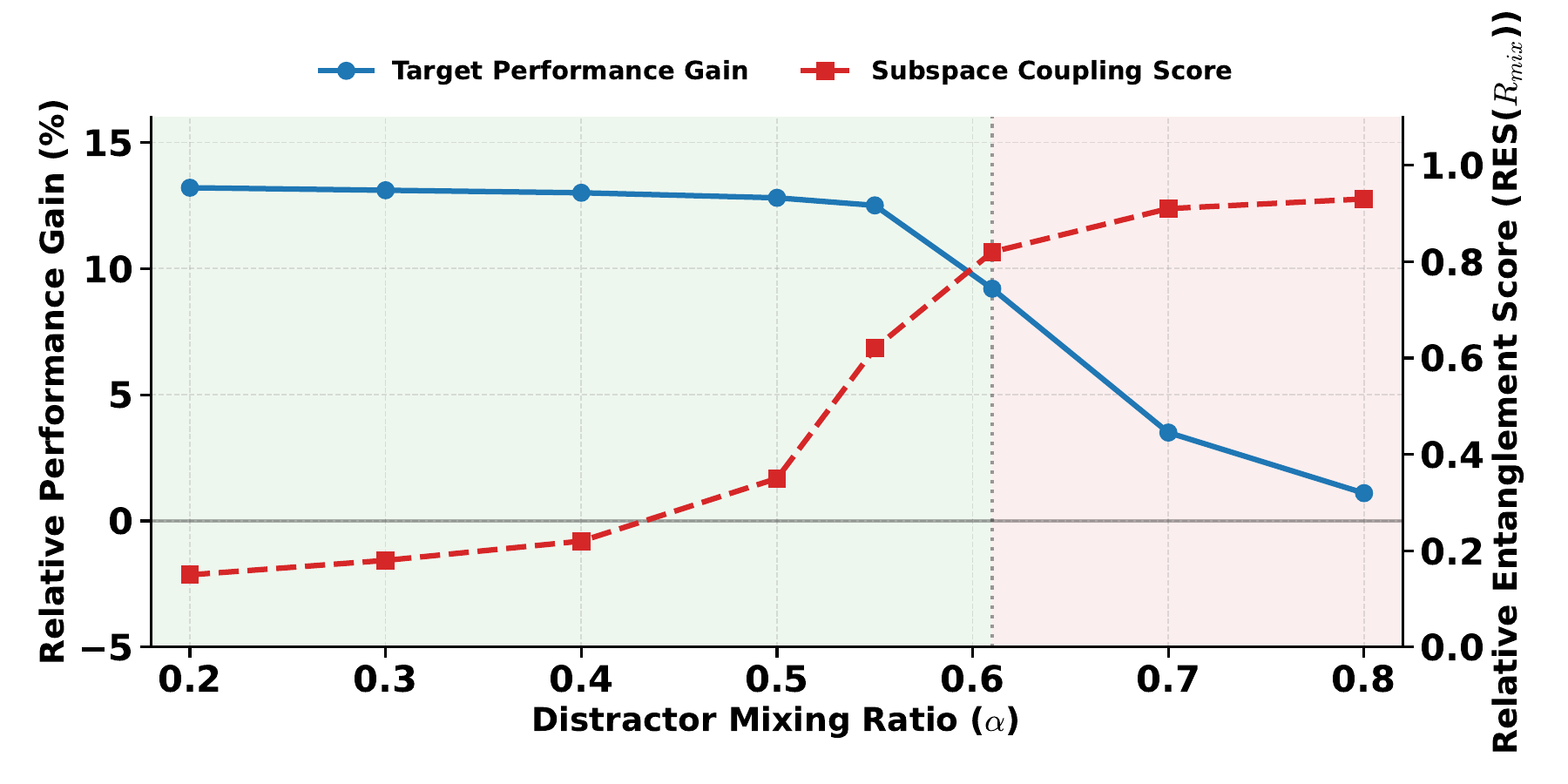}
    \caption{Impact of Data Contamination on Subspace Semantics.}
    \label{fig:data_coupling}
    \vspace{-10pt} 
\end{figure}
    performance across all the attributes, this minor overhead represents a favorable trade-off between operational efficiency and multi-attribute steering capability.

\subsection{Sensitivity to Data Coupling}
\label{sec:data_coupling_main}
To assess the robustness of MSRS against data contamination arising from attribute co-occurrence (where target samples inextricably contain distractor features), we investigate how \textit{data-level coupling} transmits to \textit{representation-level entanglement}. We simulated varying degrees of contamination by blending a pure target dataset (\textit{Helpfulness}) with a distractor dataset (\textit{Coherence}) at a mixing ratio $\alpha$. We trained steering matrices on these mixed sets and evaluated their semantic alignment using a \textbf{Relative Entanglement Score (RES)}, which measures the subspace's projection ratio between the distractor and the target oracle (detailed setup and metric definitions are provided in Appendix~\ref{sec:data_coupling_analysis}). As shown in Figure~\ref{fig:data_coupling}, MSRS exhibits a distinct non-linear robustness profile. In the \textbf{Robustness Regime} ($\alpha \le 0.55$), the method demonstrates remarkable resilience: even when distractor data constitutes the majority, the learned subspace maintains low entanglement ($\text{RES} < 0.62$) and stable steering performance ($+12.5\%$). A critical \textbf{Breakdown Point} is observed at $\alpha \approx 0.61$, where the RES surges and performance drops sharply, indicating a "tipping point" where distractor gradients begin to dominate. Notably, even under severe contamination ($\alpha=0.80$), MSRS retains positive steering efficacy ($+1.5\%$), confirming its intrinsic capability to filter data noise and extract discriminative features, validating the robustness of our subspace-based intervention.

 \begin{figure}[t]
    \centering
\includegraphics[width=0.9\linewidth,height=0.27\textheight,keepaspectratio]{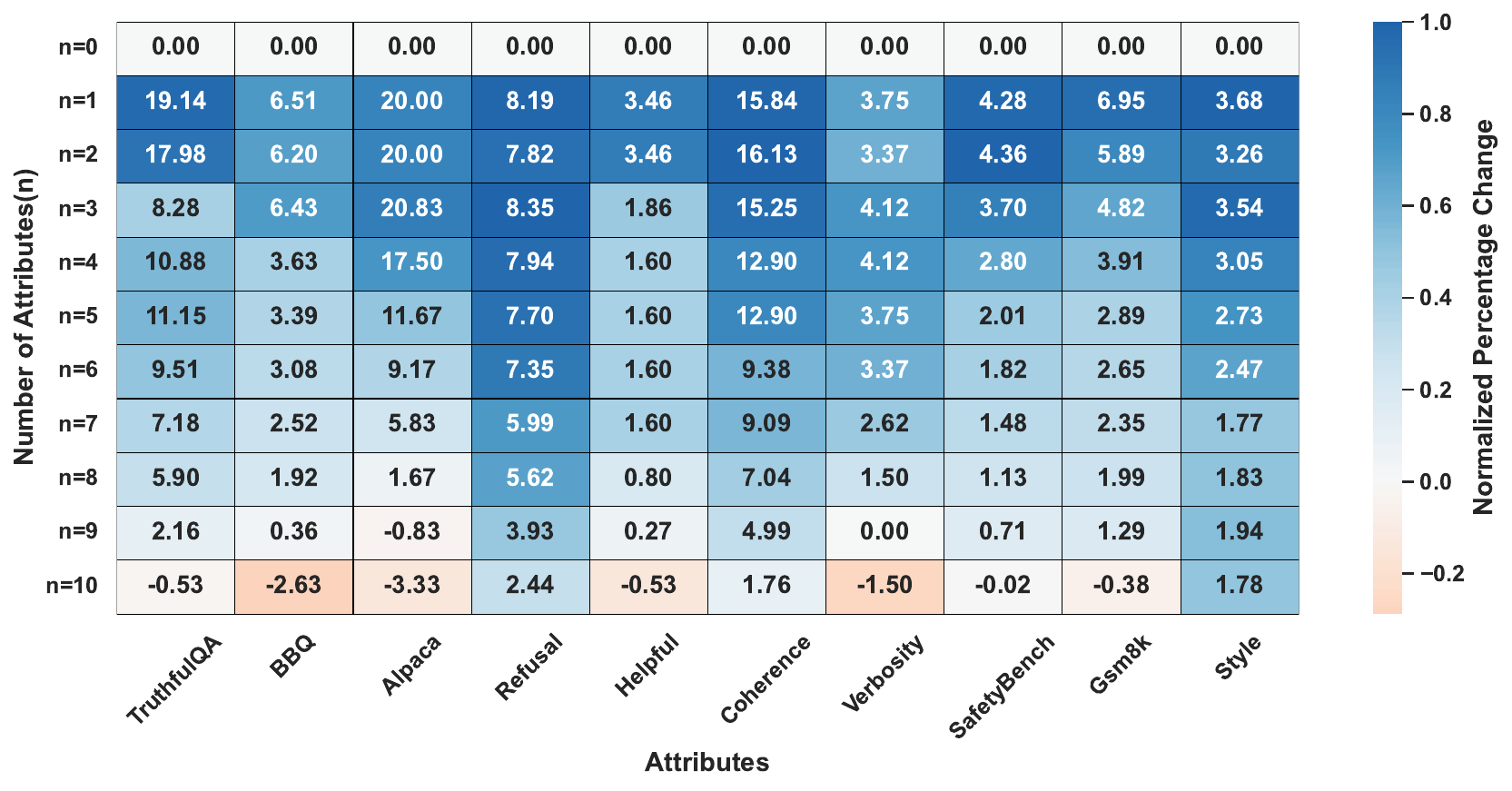}
    \caption{Normalized Relative Percentage Change (\(\Delta\%\)) in Different Attributes Compared to Baseline (n = 0).}
    \label{fig:figure5}
    \vspace{-6pt} 
\end{figure}

\begin{figure}[t]
    \centering
\includegraphics[width=0.9\linewidth,height=0.27\textheight,keepaspectratio]{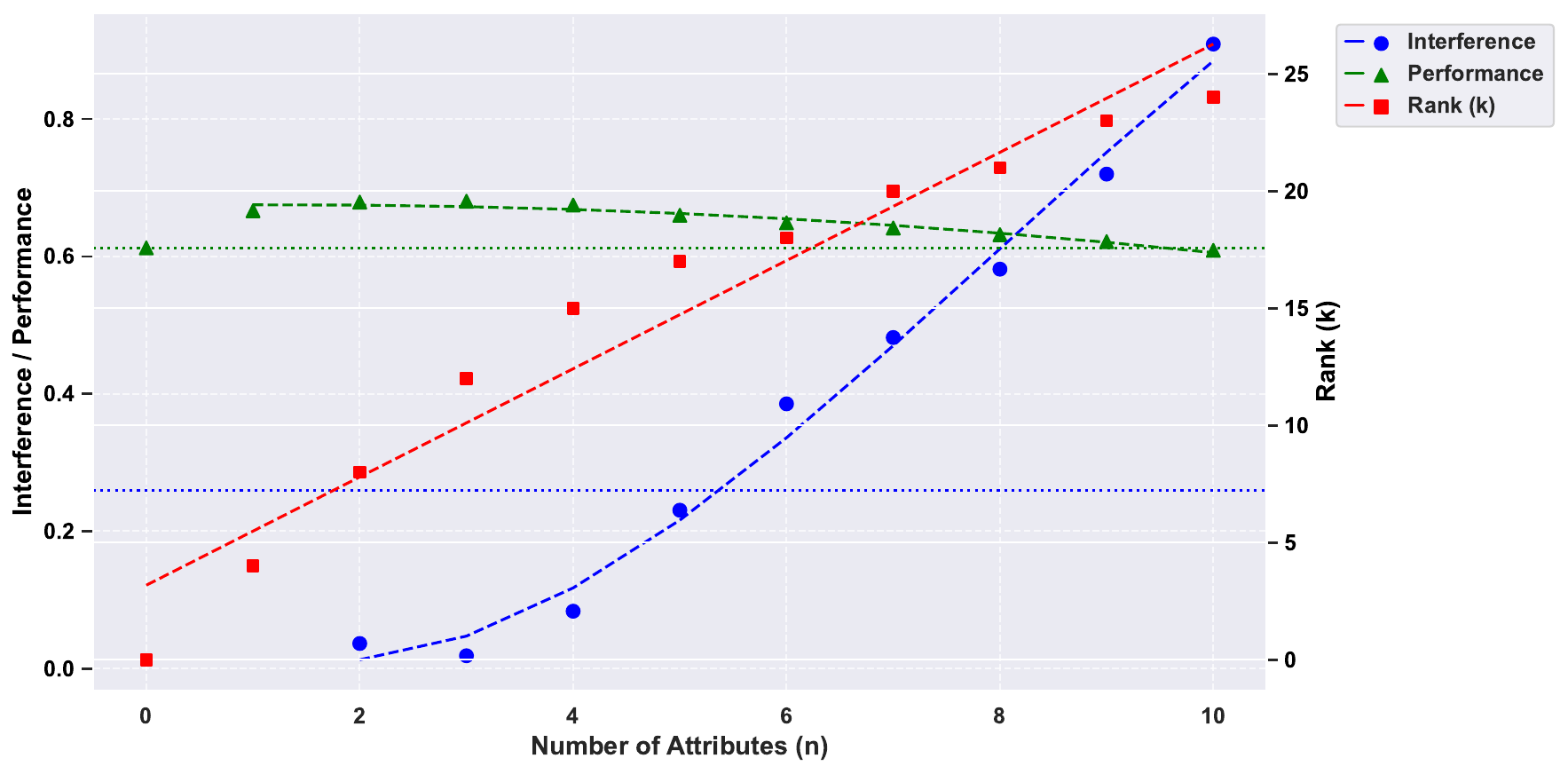}
    \caption{Interference, Performance, and Rank(k) vs. Number of Attributes(n).}
    \label{fig:figure6}
    \vspace{-12pt} 
\end{figure}

\subsection{Scalability and Interference Patterns with Growing Attribute Count}
\label{analysis:scaling}
\textbf{Investigate the relationship between the number of attributes \( n \) and performance in multi-attributes scenario.} Figure~\ref{fig:figure5} visualizes the relative performance change (\(\Delta\%\)) for each attribute \(a_i\) (\(i \in \{1, \dots, 10\}\)) across \(n \in \{0, \dots, 10\}\) attributes (attributes details are shown in Appendix~\ref{appendix:algorithm-a2}), relative to the baseline (\(n=0\)). For each attribute \(a_i\) and number of attributes \(n=j\), we construct $M$ random subsets \(S_{i,j}^m\) (\(m \in \{1, \dots, M\}\)), each defined as \(S_{i,j}^m = \{a_i\} \cup T_{i,j}^m\), where \(T_{i,j}^m \subset \mathcal{A} \setminus \{a_i\}\) is a random subset of \(j-1\) attributes sampled from the 9 attributes excluding \(a_i\), with \(|\mathcal{A}| = 10\); in our setup, \(  M=5 \). The performance of \(a_i\) in \(j\)-joint attributes is measured after joint training, and averaged as \(\text{Perf}_{a_i}(n=j) = \frac{1}{M} \sum_{m=1}^M \text{Perf}_{a_i}(S_{i,j}^m)\).
The baseline is \(\text{Perf}_{a_i}(n=0)\), and the relative performance change is given by \(\Delta_{a_i}(n=j) = \frac{\text{Perf}_{a_i}(n=j) - \text{Perf}_{a_i}(n=0)}{\text{Perf}_{a_i}(n=0)}\). The \(\Delta_{a_i}(n)\) is normalized to [0, 1] as \(\text{Norm}_{\Delta_{a_i}}(n) = \frac{\Delta_{a_i}(n)}{\max_n \Delta_{a_i}(n) - \min_n \Delta_{a_i}(n)}\). The heatmap shows \(\text{Norm}_{\Delta_{a_i}}(n)\) as color intensity (red: negative, blue: positive) with \(\Delta_{a_i}(n)\) annotated (See Table~\ref{tab:number-setting} for the performance results).  As shown in  Figure~\ref{fig:figure5}, for \(n=2-4\), most attributes show strong improvements (\(\Delta\% > 0\), deep blue), indicating effective attributes joint. At \(n=5-8\), gains weaken (lighter blue), reflecting uneven benefits. At \(n=9-10\), some attributes drop to or below baseline (\(\Delta\% \leq 0\), red/neutral), signaling performance degradation. We attribute these trends to interference in the shared subspace driven by increasing attribute counts. At low \(n\), attributes align with minimal interference. As \(n\) grows, dominant attributes (e.g., Refusal, Style) with positive \(\Delta\%\) bias the shared subspace, their steering vectors overpowering weaker attributes, which exhibit sharply reduced gains. At \(n=9-10\), intensified interference causes performance declines across more attributes.

\textbf{Performance, interference, and rank vary with number of attributes \( n \) in multi-attribute steering.} Figure~\ref{fig:figure6} illustrates the relationships between average performance (green), interference (blue), and rank \(k\) (red) with \(n\). Performance is computed as the mean over all attributes. Interference quantifies disorder in the shared subspace, manifested as consistent performance declines across attributes. It is defined as the mean normalized decline from peak performance: for each attribute \(a_j\), the decline is given by \(\text{Decline}_{a_j}(n) = \frac{\text{PeakPerf}_{a_j} - \text{Perf}_{a_j}(n)}{\text{PeakPerf}_{a_j}}\), with \(\text{PeakPerf}_{a_j} = \max_n \text{Perf}_{a_j}(n)\) as the highest performance and \(\text{Perf}_{a_j}(n)\) as the performance at \( n \). The average interference is computed as \(\text{AvgInterference}(n) = \frac{1}{m} \sum_{j=1}^m \text{Decline}_{a_j}(n)\), where \(m=10\) is the number of attributes.
\vspace{-0.5pt}
The trends approximate a quadratic fit for performance (\(R^2 = 0.9520\)), a linear fit for rank \(k\) (\(R^2 = 0.9427\)), and a cubic fit for interference (\(R^2 = 0.9903\)) (see Appendix~\ref{app:fit_parameters} for fit parameters). The green performance curve shows sustained improvements over baseline for \(n < 9\), peaking at \(n=2\) or 3 with subsequent gradual reductions in gains. The blue curve indicates accelerating interference with \(n\), while optimal performance corresponds to rank \(k=8\) to 12, consistent with Table~\ref{tab:rank-sensitivity}. These patterns suggest that at low \(n\), attributes align synergistically, enabling superior multi-attribute steering performance and demonstrating attribute scalability for \(n < 9\), corresponding \(\text{AvgInterference} < 0.6\).

\section{Conclusion}
We present Multi-Subspace Representation Steering (MSRS), a principled and effective framework for multi-attribute behavior control in LLMs. MSRS addresses key limitations of prior methods by introducing attribute-specific subspaces with SVD-guided dimensionality, a shared subspace to capture cross-attribute correlations for precise intervention. By integrating these components, MSRS mitigates steering conflicts and improves controllability across diverse attributes. Extensive experiments demonstrate that MSRS consistently outperforms existing approaches in both task-specific and general-purpose settings, offering a scalable and robust solution for reliable and aligned language generation.

\section*{Impact Statement}
This paper presents work whose goal is to advance the field of Machine
Learning. There are many potential societal consequences of our work, none of which we feel must be specifically highlighted here.

\bibliography{main}
\bibliographystyle{icml2026}

\newpage
\appendix
\onecolumn
\section{Implementation Details}
\label{appendix:algorithm}
\noindent\textbf{A.1 Datasets and Metrics}
\label{appendix:algorithm-a2}
\noindent We provide detailed descriptions of the datasets and evaluation metrics used in our experiments to assess multi-attribute steering performance.

\paragraph{TruthfulQA} TruthfulQA~\citep{lin-etal-2022-truthfulqa} evaluates a model’s ability to produce truthful and informative responses.  
We report:
\begin{itemize}
    \item \textbf{MC1 (Single-true)}: Accuracy in selecting the single correct answer (highest log-probability among 4–5 candidates).
    \item \textbf{MC2 (Multi-true)}: Normalized probability assigned to all true reference answers.
    \item \textbf{BLEURT}, \textbf{BLEU}, and \textbf{ROUGE}: Generation-level similarity scores, computed as the difference between the maximum similarity to any true answer and any false answer.
    \item \textbf{GPT-judge} and \textbf{GPT-info}: GPT-based classifiers trained to predict human ratings of truthfulness and informativeness.
\end{itemize}

\paragraph{BBQ}The Bias Benchmark for QA (BBQ)~\citep{parrish-etal-2022-bbq} measures social bias in QA outputs across nine social dimensions (e.g., race, gender). We report accuracy: whether the model selects the correct answer.

\paragraph{AlpacaEval}~\citep{alpaca_eval}: Measures instruction-following ability via win rate against a strong baseline (test-davinci-003), judged by GPT-4o. 
    
\paragraph{Sorry-Bench}~\citep{xie2025sorrybench}: Evaluates instruction refusal on harmful inputs using a fine-tuned expert model (Mistral-7B-Instruct-v0.2). We report refusal accuracy based on the model’s ability to reject malicious or unethical instructions.

\paragraph{HelpSteer} ~\citep{wang2023helpsteer} is a human-aligned benchmark for evaluating model helpfulness. Each response is rated by GPT-4o across:
\begin{itemize}
    \item \textbf{Helpfulness}: Relevance and utility of the response.
    \item \textbf{Coherence}: Logical consistency and fluency.
    \item \textbf{Verbosity}: Appropriateness of response length.
\end{itemize}
Scores range from 0 (poor) to 4 (excellent), and we report the average for each dimension.
\paragraph{SafetyBench}~\citep{zhang2023safetybench}: Measures performance on multiple safety dimensions including offensiveness, unfairness, ethics and morality, and privacy and property. This benchmark evaluates the model’s ability to generate safe and responsible responses, covering a broad range of sensitive topics. The evaluation involves assessing the degree of harmfulness in responses, and how well the model adheres to ethical and privacy considerations.

\paragraph{GSM8k}~\citep{cobbe2021gsm8k}: Evaluates reasoning ability by testing the model on a variety of mathematical word problems. The GSM8k benchmark is designed to assess the model's ability to perform multi-step reasoning and inference, as well as its ability to handle mathematical tasks at a high level of accuracy.

\paragraph{Human-Style-Answers}: A benchmark designed to evaluate the style of human-like responses generated by AI models. It focuses on generating natural, contextually appropriate, and coherent responses. The evaluation includes assessing the fluency, tone, and style of the model’s responses, ensuring they are aligned with typical human responses. Details of the benchmark can be found at \url{https://huggingface.co/datasets/innova-ai/Human-Style-Answers}.

\paragraph{General Benchmarks.}
To verify that steering does not impair general capabilities, we evaluate on standard NLP tasks:
\begin{itemize}
    \item \textbf{HellaSwag}~\citep{zellers-etal-2019-hellaswag}: commonsense inference; metric: \textit{accuracy}.
    \item \textbf{RACE}~\citep{lai-etal-2017-race}: reading comprehension; metric: \textit{accuracy}.
    \item \textbf{OpenBookQA}~\citep{mihaylov-etal-2018-suit}: elementary science QA; metric: \textit{accuracy}.  
\end{itemize}

\noindent \textbf{MMLU}~\citep{hendrycks2020measuring} (Massive Multitask Language Understanding) is a challenging benchmark designed to evaluate a model’s world knowledge and problem-solving ability under zero-shot and few-shot settings. It comprises 15,908 multiple-choice questions spanning 57 diverse subjects, including STEM, humanities, social sciences, and professional disciplines such as law and ethics. The tasks vary in difficulty from elementary to advanced levels, making MMLU an ideal benchmark for identifying model weaknesses across both general and specialized domains.

Each subject contains at least 100 test questions, exceeding the length of most human exams. The dataset is split into a few-shot development set (5 questions per subject), a validation set (1,540 questions), and a test set (14,079 questions). The evaluation metric is \textit{average accuracy} across all subjects.

\noindent \textbf{GLUE}~\citep{wang2018glue} (General Language Understanding Evaluation) is a widely used benchmark for evaluating general-purpose language understanding. It consists of nine diverse NLP tasks that span a range of linguistic phenomena, including sentiment analysis, paraphrase detection, textual entailment, and question answering. These tasks collectively assess a model’s ability to perform natural language understanding in varied contexts.

The included tasks are:

\textbf{MNLI} (Multi-Genre Natural Language Inference): Predict entailment, contradiction, or neutrality between premise and hypothesis across multiple domains.

\textbf{QNLI} (Question Natural Language Inference): Convert question answering into an entailment task.

\textbf{QQP} (Quora Question Pairs): Detect if two questions from Quora have the same meaning.

\textbf{SST-2} (Stanford Sentiment Treebank): Classify sentiment in movie reviews as positive or negative.

\textbf{CoLA} (Corpus of Linguistic Acceptability): Judge grammatical acceptability of a sentence.

\textbf{STS-B} (Semantic Textual Similarity Benchmark): Score sentence pairs on semantic similarity.

\textbf{MRPC} (Microsoft Research Paraphrase Corpus): Determine if two sentences are paraphrases.

\textbf{RTE} (Recognizing Textual Entailment): Binary entailment classification from multiple datasets.

\textbf{WNLI} (Winograd NLI): Resolve coreference in complex pronoun cases.

\begin{table*}[ht]
\caption{MMLU per-task performance on different methods. The best result is highlighted in bold, and the second-best is underlined.}
\centering
\resizebox{\textwidth}{!}{
\begin{tabular}{lccccccccccccccccc}
\hline\hline
\textbf{Method} & Math & Health & Physics & Business & Biology & Chemistry & CS & Economics & Eng. & Philosophy & Other & History & Geog. & Politics & Psych. & Culture & Law \\
\hline
LLaMA3-8B-Instruct & 0.430 & \underline{0.697} & 0.533 & 0.819 & 0.791 & 0.502 & 0.629 & 0.675 & 0.641 & 0.567 & 0.694 & 0.778 & 0.848 & 0.796 & 0.764 & 0.816 & 0.516 \\
+ ITI              & 0.371 & 0.597 & 0.450 & 0.703 & 0.685 & 0.472 & 0.476 & 0.566 & 0.517 & 0.486 & 0.616 & 0.624 & 0.732 & 0.667 & 0.653 & 0.738 & 0.393 \\
+ CAA              & \underline{0.436} & 0.695 & 0.545 & 0.828 & 0.786 & 0.488 & \underline{0.638} & \textbf{0.675} & \textbf{0.676} & 0.566 & 0.689 & 0.772 & \underline{0.843} & 0.787 & 0.768 & \textbf{0.852} & 0.513 \\
+ Ours             & \textbf{0.567} & \textbf{0.714} & \textbf{0.642} & \textbf{0.854} & \textbf{0.846} & \textbf{0.578} & \textbf{0.648} & \underline{0.759} & \underline{0.759} & \textbf{0.617} & \textbf{0.761} & \textbf{0.809} & \textbf{0.889} & \textbf{0.810} & \textbf{0.814} & 0.850 & \textbf{0.563} \\
Qwen2-7B-Instruct  & \underline{0.567} & 0.712 & 0.630 & \textbf{0.863} & 0.844 & 0.574 & 0.650 & 0.757 & 0.738 & 0.594 & 0.754 & 0.801 & 0.874 & 0.813 & 0.803 & 0.825 & 0.557 \\
+ CAA              & \textbf{0.568} & \underline{0.714} & \underline{0.639} & 0.854 & \textbf{0.850} & 0.578 & \textbf{0.653} & \textbf{0.760} & \underline{0.759} & \underline{0.613} & \underline{0.758} & \textbf{0.815} & \underline{0.884} & \underline{0.815} & \underline{0.813} & \textbf{0.852} & 0.561 \\
+ Ours             & 0.567 & \textbf{0.715} & \textbf{0.642} & \underline{0.856} & \underline{0.846} & \textbf{0.578} & 0.648 & \underline{0.759} & \textbf{0.762} & \textbf{0.617} & \textbf{0.761} & \underline{0.809} & \textbf{0.889} & 0.810 & \textbf{0.814} & \underline{0.851} & \textbf{0.563} \\
Mistral-7b-v0.3    & \underline{0.387} & \underline{0.658} & \underline{0.475} & \underline{0.766} & 0.742 & \textbf{0.492} & \underline{0.587} & 0.545 & \textbf{0.614} & \underline{0.567} & \underline{0.682} & 0.761 & \textbf{0.773} & \textbf{0.776} & \underline{0.737} & 0.771 & \underline{0.503} \\
+ CAA             & \textbf{0.388} & 0.659 & 0.473 & 0.765 & \underline{0.742} & \underline{0.482} & \textbf{0.595} & \textbf{0.585} & \underline{0.600} & \textbf{0.570} & \textbf{0.685} & \underline{0.760} & \underline{0.771} & 0.768 & 0.729 & \underline{0.774} & 0.497 \\
+ Ours    & 0.365 & \textbf{0.687} & \textbf{0.477} & \textbf{0.769} & \textbf{0.687} & 0.429 & 0.534 & \underline{0.547} & 0.503 & \underline{0.523} & 0.547 & 0.703 & 0.737 & \underline{0.698} & \textbf{0.790} & \textbf{0.787} & \textbf{0.532} \\          
\hline\hline
\end{tabular}
}
\label{tab:mmlu-subject}
\end{table*}

\begin{table}[ht]
\caption{Model Performance for Different Numbers of Attribute Settings.}
\centering
\resizebox{\textwidth}{!}{
\begin{tabular}{ccccccccccc}
\toprule
\textbf{r (n)} & \textbf{truthfulqa}~($\uparrow$) & \textbf{bbq}~($\uparrow$) & \textbf{alpaca}~($\uparrow$) & \textbf{refusal}~($\uparrow$) & \textbf{helpful}~($\uparrow$) & \textbf{coherence}~($\uparrow$) & \textbf{verbosity}~($\downarrow$) & \textbf{SafetyBench}~($\uparrow$) & \textbf{gsm8k}~($\uparrow$) & \textbf{style}~($\uparrow$) \\
\midrule
0 & 36.55\% & 0.6081 & 0.12 & 0.4911 & 3.76 & 3.41 & 2.33 & 62.13 & 70.32 & 75.35 \\
4 (n=1) & 43.54\% & 0.6476 & 0.35 & 0.5313 & 3.89 & 3.91 & 2.21 & 64.79 & 75.21 & 77.82 \\
8 (n=2) & 43.12\% & 0.6457 & 0.36 & 0.5295 & 3.89 & 3.96 & 2.24 & 64.84 & 74.46 & 77.81 \\
12 (n=3) & 39.57\% & 0.6471 & 0.37 & 0.5321 & 3.83 & 3.93 & 2.22 & 64.43 & 73.71 & 78.02 \\
15 (n=4) & 40.52\% & 0.6301 & 0.33 & 0.5301 & 3.82 & 3.85 & 2.22 & 63.87 & 73.07 & 77.65 \\
17 (n=5) & 38.74\% & 0.6246 & 0.26 & 0.5179 & 3.82 & 3.85 & 2.23 & 63.38 & 72.35 & 77.41 \\
18 (n=6) & 38.52\% & 0.6237 & 0.23 & 0.5172 & 3.82 & 3.73 & 2.29 & 66.26 & 72.18 & 77.21 \\
20 (n=7) & 38.47\% & 0.6233 & 0.14 & 0.5105 & 3.82 & 3.72 & 2.32 & 62.51 & 71.97 & 76.68 \\
21 (n=8) & 37.87\% & 0.6197 & 0.12 & 0.5087 & 3.79 & 3.60 & 2.35 & 62.30 & 71.72 & 76.73 \\
23 (n=9) & 37.68\% & 0.6102 & 0.11 & 0.5091 & 3.79 & 3.58 & 2.36 & 62.28 & 71.73 & 76.81 \\
24 (n=10) & 36.88\% & 0.5920 & 0.08 & 0.5031 & 3.79 & 3.47 & 2.37 & 62.12 & 70.05 & 76.69 \\
\bottomrule
\end{tabular}
}
\label{tab:number-setting}
\end{table}

\subsection{Performance on MMLU  Benchmark}
\label{appendix:algorithm-MMLU}
This section provides a comprehensive evaluation of MSRS on the Massive Multitask Language Understanding (MMLU) benchmark, assessing its ability to maintain and enhance general language capabilities across diverse domains. The MMLU benchmark, comprising 57 tasks across 17 subjects, serves as a rigorous testbed for evaluating model robustness beyond attribute-specific steering. We compare MSRS against baseline ITI and CAA on three base models: LLaMA3-8B-Instruct, Qwen2-7B-Instruct, and Mistral-7b-v0.3.  Table~\ref{tab:mmlu-subject} summarizes the per-subject performance of MSRS and baselines on the MMLU benchmark. The results reveal distinct patterns in how each method impacts general capabilities, with MSRS demonstrating superior consistency and enhancement over baselines.

\subsection{Performance on GLUE Benchmark}
\label{subsec:glue-performance}

In this section, we assess the performance of MSRS on the GLUE benchmark, a widely-used suite of tasks designed to evaluate natural language understanding capabilities. We evaluate on GLUE tasks including SST-2 (sentiment analysis), STS-B (semantic similarity), QNLI (question-answering), CoLA (linguistic acceptability), QQP (paraphrase detection), and RTE (textual entailment). We compare MSRS against two baseline methods: CAA and ReFT, across three base models: LLaMA3-8B-Instruct, Qwen2-7B-Instruct, and Mistral-7B-v0.3. Table~\ref{tab:glue-results} provides a comprehensive summary of the performance across all methods and models on the GLUE benchmark. MSRS demonstrates consistent improvements over the base models and often outperforms the baseline methods, showcasing its robustness across diverse linguistic tasks.

\begin{table}[ht]
\caption{Performance on GLUE benchmark tasks with different methods. The best result is highlighted in bold, and
the second-best is underlined. }
\centering
\resizebox{0.7\linewidth}{!}{
\begin{tabular}{lccccccc}
\hline\hline
\textbf{Model / Method} & SST-2 & STS-B & QNLI & CoLA & QQP & RTE & Avg. \\
\hline
LLaMA3-8B-Inst.   & 0.9471 & 0.5266 & 0.7208 & 0.8279 & \underline{0.6426} & \underline{0.6897} & 0.7257 \\
+ CAA            & \underline{0.9641} & 0.5743 & 0.7571 & \textbf{0.8317} & 0.6336 & 0.6701 & 0.7384 \\
+ ReFT           & 0.9585 & \underline{0.6662} & \underline{0.8018} & 0.8185 & 0.6385 & 0.6577 & \underline{0.7569} \\
+ Ours           & \textbf{0.9799} & \textbf{0.6400} & \textbf{0.8097} & \underline{0.8289} & \textbf{0.6501} & \textbf{0.6912} & \textbf{0.7578} \\
\hline
Qwen2-7B-Inst.    & \underline{0.9231} & \textbf{0.7821} & 0.8228 & 0.7500 & 0.8157 & \textbf{0.8602} & 0.8256 \\
+ CAA            & \textbf{0.9601} & 0.6342 & 0.8070 & \textbf{0.8317} & 0.6336 & 0.6701 & 0.7701 \\
+ ReFT           & 0.8750 & \underline{0.7404} & \underline{0.8637} & 0.7931 & \underline{0.8227} & 0.8498 & \underline{0.8300} \\
+ Ours           & 0.8850 & 0.7311 & \textbf{0.8675} & \underline{0.8210} & \textbf{0.8672} & \underline{0.8577}  & \textbf{0.8322} \\
\hline
Mistral-7B-v0.3   & \underline{0.8625} & \underline{0.8313} & 0.4941 & \underline{0.7731} & \underline{0.7695} & 0.3548 & 0.6808 \\
+ CAA            & \textbf{0.8671} & \textbf{0.8357} & 0.3225 & 0.8106 & \textbf{0.7727} & 0.3225 & 0.6551 \\
+ ReFT           & 0.8249 & 0.7524 & \underline{0.5401} & \underline{0.8208} & 0.6398 & \underline{0.5413} & \underline{0.6927} \\
+ Ours           & 0.8372 & 0.7972 & \textbf{0.5712} & \textbf{0.8452} & 0.6102 & \textbf{0.6153} & \textbf{0.7066} \\
\hline\hline
\end{tabular}
}

\label{tab:glue-results}
\end{table}

\section{Computational Complexity and Runtime Analysis}
\label{app:complexity}
All experiments are conducted on a single NVIDIA V100 32 GiB GPU. No specialized optimizations (e.g., randomized SVD or CPU offloading) are used.
We analyze the computational cost of the subspace extraction phases:

\subsection{Offline Subspace Extraction}
This phase is performed once per model and attribute combination.

\textbf{Shared Subspace Extraction.} We first forward each balanced dataset \(\mathcal{D}_i\) (\(|\mathcal{D}_i| \leq 1000\)) through the frozen LLM and compute the mean activation vector \(\tau_i \in \mathbb{R}^d\) (\(d=4096\)) from the last few tokens. These vectors are concatenated into:
\[
\tau_c = [\tau_1 \,|\, \tau_2 \,|\, \dots \,|\, \tau_n] \in \mathbb{R}^{d \times n}.
\]
A single full SVD is performed on \(\tau_c\) with complexity \(\mathcal{O}(d n^2)\).

\textbf{Iterative Attribute-Specific Subspace Extraction.} For each attribute \(i\), we form the per-sample activation matrix
\[
H_i^{(i)} = [h_{i,1}^l, \dots, h_{i,|\mathcal{D}_i|}^l] \in \mathbb{R}^{d \times |\mathcal{D}_i|}, \quad |\mathcal{D}_i| \leq 1000.
\]
To ensure orthogonality among all learned subspaces, we iteratively subtract the projections of the shared subspace and all previously extracted attribute subspaces. The residual is computed as:
\begin{equation}
   H_{\text{res}}^{(i)} = H_i^{(i)} - B_{\text{shared}}^\top B_{\text{shared}} H_i^{(i)} - \sum_{k=1}^{i-1} B_k^\top B_k H_i^{(i)},
\end{equation}
where $B_{\text{shared}}$ and $B_k$ are the orthonormal bases for the shared and prior attribute subspaces ($k < i$), respectively. We then perform full SVD on each \(H_{\text{res}}^{(i)}\) independently.
Theoretical complexity per attribute: \(\mathcal{O}(d \cdot |\mathcal{D}_i|^2) \approx \mathcal{O}(4{,}096 \times 10^6) \approx 4 \times 10^9\) FLOPs in the worst case.

\textbf{Measured Wall-Clock Time}

\begin{table}[h]
\centering
\caption{Offline subspace extraction time on Llama-3-8B for $n$ attributes and $N$ samples per attribute.}
\resizebox{\textwidth}{!}{
\begin{tabular}{lccc}
\toprule
Phase & Matrix size & Time per run (seconds) & Total Time (seconds) \\
\midrule
Activation extraction & - & 0.29 (per sample) & $N \times n \times 0.29$ \\
Shared subspace & $4096 \times n$ & - & 0.18 \\
Per-attribute subspace & $4096 \times \leq\!1000$ each & 3.42 (per attribute) & $3.42 \times n$ \\
\midrule
\textbf{Full offline phase} & — & — & $0.18 + (3.42 \times n) + (N \times n \times 0.29)$ \\
\bottomrule
\end{tabular}
}
\label{tab:offline-time}
\end{table}

Realistic estimate for the setting (\(n=3\), with \(N=1000\) samples per attribute):
\[
\text{Total time} \approx 0.18 + (3.42 \times 3) + (1000 \times 3 \times 0.29) \approx 0.18 + 10.26 + 870 = 880.44 \text{ seconds}
\]
This cost is incurred only once per model and attribute set. The vast majority of this time is spent on forward passes to collect activations. The resulting subspace bases \(B_{\text{shared}}\) and \(\{B_i\}_{i=1}^n\) are saved and directly loaded by all subsequent runs, introducing zero additional overhead during online training or inference.

\subsection{Online Training Cost}
Following the subspace extraction, we proceed to the online training phase where the steering parameters (e.g., the mask network $m_\theta$ and projection matrices) are optimized.

\textbf{Experimental Setup.} We train MSRS using the same hardware (single V100 32GB) with the following hyperparameters: rank $r=8$, batch size $B=4$, and a dataset size of $N=500$ samples per attribute. The model is trained for $E=10$ epochs.

\textbf{Complexity \& Overhead.} The computational graph of MSRS is nearly identical to standard ReFT (Low-rank Linear Subspace ReFT), as the additional mask prediction network is a lightweight MLP ($d \to n$) with negligible FLOPs compared to the Transformer backbone's forward/backward pass.

\textbf{Measured Training Time.}
Table~\ref{tab:online-time} summarizes the training throughput. With gradient checkpointing enabled, the average time per training step is approximately 0.45 seconds.

\begin{table}[h]
\centering
\caption{Online training cost comparison on Llama-3-8B ($n=3$ attributes, $N=500$ per attribute).}
\resizebox{0.9\textwidth}{!}{
\begin{tabular}{lccccc}
\toprule
Method & Params & Time per Step & Total Steps & \multicolumn{2}{c}{Total Training Time} \\
& & ($B=4$) & ($E=10$) & (Seconds) & (GPU Hours) \\
\midrule
LoReFT (Baseline) & $d \times r$ & 0.42s & 3,750 & $\approx 1,575$s & $\approx 0.44$h \\
\textbf{MSRS (Ours)} & $d \times r + \text{Mask}$ & 0.45s & 3,750 & $\approx 1,688$s & $\approx \mathbf{0.47}$h \\
\bottomrule
\end{tabular}
}
\label{tab:online-time}
\end{table}

\noindent For this setting ($n=3$, total samples $1,500$), the total training budget is:
\[
\text{Cost} = E \times \frac{N_{\text{total}}}{B} \times T_{\text{step}} \approx 10 \times \frac{1500}{4} \times 0.45s \approx 1688 \text{ seconds} \approx \mathbf{0.47 \text{ GPU Hours}}.
\]
This demonstrates that MSRS requires less than 0.5 GPU hours to converge, making it highly efficient for rapid experimentation and deployment.

\section{Adaptive Subspace Selecting }
\label{appendix:C}
This section presents an in-depth evaluation of the adaptive subspace selecting mechanism in MSRS. We assess the effectiveness of three subspace training strategies. (1) Same Space, where all attributes share a single subspace; (2) \(\text{MSRS}_{\text{Attribute}}\), where a mask network adaptively weights attribute-specific subspaces; and (3) \(\text{MSRS}_{\text{Rank}}\), where the mask network weights individual low-rank dimensions. Table~\ref{tab:ablation-space} summarizes the performance of these strategies on TruthfulQA, BBQ, Alpaca, Refusal, and HelpSteer datasets, evaluated with LLaMA3-8B-Instruct, Qwen2-7B-Instruct, and Mistral-7B-v0.3 models. The analysis highlights the limitations of the Same Space approach and the advantages of adaptive subspace mechanisms.

\begin{table*}[t!]
\caption{Comparison of steering subspace training strategies across datasets. The best result is highlighted in bold.}
\centering
\resizebox{\textwidth}{!}{%
\begin{tabular}{lcccccccccc}
\hline\hline
\textbf{Method} & \multicolumn{4}{c}{\textbf{TruthfulQA}} & \textbf{BBQ} & \textbf{Alpaca} & \textbf{Refusal} & \multicolumn{3}{c}{\textbf{HelpSteer}} \\
& MC1 & MC2 & BLEU & BLEURT & Acc & Win Rate & Sorry-Bench & Help. & Coher. & Verb. \\
\hline
LLaMA3-8B-Instruct & 29.58 & 48.43 & 49.63 & 57.88 & 0.608 & 0.12 & 0.491 & 3.78 & 3.91 & 2.33 \\
Same Space         & 29.58 & 49.51 & 52.08 & 64.06 & 0.637 & 0.30 & 0.451 & 3.87 & 3.89 & 2.38 \\
\( MSRS_{attribute} \) & 32.52 & 52.55 & \textbf{52.57} & 68.46 & 0.627 & \textbf{0.36} & \textbf{0.529} & 3.88 & \textbf{3.96} & \textbf{2.24} \\
\( MSRS_{rank} \)  & \textbf{33.50} & \textbf{52.74} & 52.32 & \textbf{66.75} & \textbf{0.646} & 0.35 & 0.527 & \textbf{3.89} & 3.95 & 2.28 \\
\hline
Qwen2-7B-Instruct & 26.38 & 45.41 & 49.63 & 65.28 & 0.638 & 0.12 & 0.384 & 3.51 & 3.83 & 2.28 \\
Same Space         & 29.83 & 48.69 & 52.57 & 71.15 & 0.637 & 0.434 & 0.422 & 3.63 & 3.78 & 2.38 \\
\( MSRS_{attribute} \) & \textbf{34.72} & \textbf{53.27} & \textbf{53.10} & \textbf{74.90} & \textbf{0.642} & \textbf{0.451} & \textbf{0.446} & 3.64 & 3.81 & 2.20 \\
\( MSRS_{rank} \)  & 26.41 & 47.65 & 49.88 & 64.30 & 0.635 & 0.442 & 0.439 & \textbf{3.76} & \textbf{3.82} & \textbf{2.17} \\
\hline
Mistral-7B-v0.3  & 18.83 & 36.54 & 41.56 & 54.52 & 0.614 & 0.14 & 0.632 & 3.75 & 3.92 & 2.36 \\
Same Space         & 30.32 & 49.69 & 49.39 &  66.01 & 0.615 & 0.33 &0.669 & 3.82 & 3.85 & 2.33 \\
\( MSRS_{attribute} \) & \textbf{30.07} & \textbf{52.62} & \textbf{50.61} &  \textbf{71.39} & \textbf{0.644} &0.38 &\textbf{0.693}& \textbf{3.82} & \textbf{3.93} & 2.27 \\
\( MSRS_{rank} \)  & 28.36 & 49.94 & 47.19 &  69.19 &0.631 &\textbf{0.39} &0.673  & 3.76 & 3.87 & \textbf{2.26} \\
\hline\hline
\end{tabular}
}

\label{tab:ablation-space}
\end{table*}

\section{Intervention Position Experiments}
\label{appendix:D}
\begin{table*}[ht]
\caption{Comparison of Last token vs. Important token intervention. The best result is highlighted in bold.}
\centering
\resizebox{\textwidth}{!}{%
\begin{tabular}{lccccccccccccccccccc}
\hline\hline
\textbf{Method / Position} & \multicolumn{7}{c}{\textbf{TruthfulQA}} & {\textbf{BBQ}} & \textbf{Alpaca} & \textbf{Refusal} & \multicolumn{3}{c}{\textbf{HelpSteer}}   \\
\hline
 & MC1 & MC2 & BLEU & rouge1 & BLEURT  & Judge & Info & acc  & Win Rate & Sorry-Bench & Help. & Coher. & Verb.  \\
\hline
\textbf{LLaMA2-7B} & & & & & & & & & & & & \\
Last Token & 26.41 & 42.88 & 48.66 & 46.45 & 58.19 & \textbf{31.05} & 67.48 & 0.631 & 0.12 & 0.579 & 2.70 & 2.68 & 2.73\\
Important Token  & \textbf{29.10} & \textbf{48.60} & \textbf{49.88} & \textbf{50.37} & \textbf{60.15} & 28.85 & \textbf{75.79} & \textbf{0.644} & \textbf{0.13} & \textbf{0.583} & \textbf{3.12} & \textbf{3.06} & \textbf{2.47} \\
\hline
\textbf{LLaMA3-8B-Instruct} & & & & & & & & & & & & \\
Last Token & 33.50 & 52.74 & 52.32 & 56.48 & 66.75 & 24.69 & 76.77 & 0.646 & \textbf{0.36} & \textbf{0.529} & \textbf{3.88} & \textbf{3.96} & 2.24 \\
Important Token  & \textbf{33.71} & \textbf{56.32} & \textbf{52.71} & \textbf{58.22} & \textbf{67.51} & \textbf{29.21} & 78.13 & 0.655 & 0.32 & 0.511 & 3.85 & 3.95 & \textbf{1.99} \\
\hline
\textbf{Qwen2-7B-Instruct} & & & & & & & & & & & & \\
Last Token & 34.72 & 53.27 & 51.10 & 55.50 & 70.90 & 28.85 & \textbf{85.57} & 0.6421 & \textbf{0.45} & 0.446 & \textbf{3.70} & 3.82 & 2.17 \\
Important Token  & \textbf{36.12} & \textbf{55.63} & \textbf{52.17} & \textbf{57.25} & \textbf{70.93} & \textbf{31.41} & 84.09 & \textbf{0.657} & 0.42 & \textbf{0.448} & 3.69 & \textbf{3.83} & \textbf{2.06} \\
\hline\hline
\end{tabular}
}

\label{tab:performance-comparison-all}
\end{table*}

This section provides a detailed evaluation of the dynamic intervention position selection mechanism. We compare two intervention strategies: (1) \textit{Last Token}, where steering is applied to the final token in the sequence, and (2) \textit{Important Token}, where steering is dynamically applied to the token most relevant to the target attribute, as identified by subspace projections. The experimental results, presented in Table~\ref{tab:performance-comparison-all}, span multiple datasets (TruthfulQA, BBQ, Alpaca, Refusal, and HelpSteer) and models (LLaMA2-7B, LLaMA3-8B-Instruct, and Qwen2-7B-Instruct). Below, we analyze the effectiveness of the \textit{Important Token} strategy and its advantages over the \textit{Last Token} baseline.


\section{Steering Layer Selection }
\label{appendix:E}
\begin{table*}[ht]
\caption{Performance of interventions at different layers. The best result is highlighted in bold.}
\centering
\resizebox{\textwidth}{!}{%
\begin{tabular}{lcccccccccccccccccc}
\hline\hline
\textbf{Model / Layer} & \multicolumn{7}{c}{\textbf{TruthfulQA}} & \textbf{BBQ} & \textbf{Alpaca} & \textbf{Refusal} & \multicolumn{3}{c}{\textbf{HelpSteer}} \\
 & MC1 & MC2 & BLEU & ROUGE1 & BLEURT & Judge & Info & Acc & Win Rate & Sorry-Bench & Help. & Coher. & Verb. \\
\hline
\textbf{LLaMA3-8B-Instruct} \\
3  & 29.83 & 48.95 & 52.08 & 55.26 & 65.28 & 32.27 & \textbf{91.20} & 0.631 & 0.27 & 0.483 & 3.78 & 3.81 & 2.37 \\
9  & 33.01 & \textbf{56.39} & 49.88 & 55.26 & \textbf{68.22} & 22.00 & 82.89 & 0.620 & 0.34 & 0.517 & 3.79 & 3.85 & 2.28 \\
15 & \textbf{33.50} & 52.74 & \textbf{52.32} & \textbf{56.48} & 66.75 & 24.69 & 76.77 & \textbf{0.646} & \textbf{0.36} & \textbf{0.529} & \textbf{3.88} & \textbf{3.96} & \textbf{2.24} \\
21 & 27.94 & 47.21 & 52.18 & 55.02 & 65.45 & \textbf{26.31} & 72.26 & 0.632 & 0.31 & 0.462 & 3.87 & 3.87 & 2.36 \\
27 & 26.86 & 49.24 & 49.83 & 53.12 & 63.77 & 25.23 & 71.85 & 0.622 & 0.18 & 0.491 & 3.68 & 3.91 & 2.33 \\
\hline
\textbf{Qwen2-7B-Instruct} \\
3  & 36.32 & 44.39 & 47.79 & 54.55 & 66.81 & 20.82 & 75.77 & 0.605 & 0.32 & 0.419 & 3.72 & 3.81 & 2.34 \\
9 & 34.72 & \textbf{53.27} & \textbf{53.10} & 55.50 & \textbf{74.90} & 28.85 & \textbf{90.95} & \textbf{0.642} & \textbf{0.45} & 0.446 & \textbf{3.76} & 3.82 & \textbf{2.17} \\
15  & \textbf{36.41} & 47.65 & 49.41 & \textbf{58.03} & 63.80 & 25.43 & 87.31 & 0.605 & 0.44 & \textbf{0.449} & 3.64 & \textbf{3.84} & 2.25 \\
21 & 33.67 & 50.79 & 51.87 & 53.28 & 73.41 & 31.41 & 77.18 & 0.631 & 0.38 & 0.413 & 3.63 & 3.76 & 2.27 \\
27 & 24.71 & 40.44 & 49.83 & 53.30 & 69.47 & \textbf{36.62} & 74.76 & 0.614 & 0.16 & 0.368 & 3.61 & 3.61 & 2.24 \\
\hline
\textbf{Mistral-7B-v0.3} \\
3  & 27.06 & 46.94 & 48.09 & 54.26 & 68.49 & 44.03 & 67.95 & 0.631 & 0.36 & 0.622 & \textbf{3.86} & 3.71 & 2.32 \\
9  & 30.82 & 49.06 & 47.79 & 53.25 & 69.01 & \textbf{51.07} & 79.53 & 0.624 & 0.32 & 0.679 & 3.82 & 3.87 & 2.31 \\
15 & \textbf{31.32} & 51.69 & 48.59 & 57.58 & 63.87 & 50.37 & 78.57 & 0.619 & 0.36 & 0.669 & 3.76 & 3.88 & \textbf{2.23} \\
21 & 30.07 & \textbf{52.62} & \textbf{50.61} & \textbf{57.95} & \textbf{71.39} & 45.70 & 80.44 & \textbf{0.644} & \textbf{0.38} & \textbf{0.693} & 3.82 & \textbf{3.93} & 2.27 \\
27 & 24.83 & 36.31 & 45.76 & 48.64 & 53.52 & 44.86 & \textbf{84.03} & 0.614 & 0.25 & 0.619 & 3.71 & 3.72 & 2.35 \\
\hline\hline
\end{tabular}
}

\label{tab:performance-metrics}
\end{table*}

This section elaborates on the layer-wise ablation study conducted to identify the optimal transformer layer for injecting steering vectors in MSRS. We assess model performance by applying interventions at specific layers (\{3, 9, 15, 21, 27\}) across multiple datasets and models. The analysis highlights the sensitivity of steering effectiveness to layer selection and underscores the importance of optimizing this parameter. Table~\ref{tab:performance-metrics} summarizes the performance metrics for interventions at different layers, evaluated on TruthfulQA, BBQ, Alpaca, Refusal, and HelpSteer datasets using LLaMA3-8B-Instruct, Qwen2-7B-Instruct, and Mistral-7B-v0.3 models.

\textbf{Layer Selection via Grid Search}  
To determine the optimal intervention layer for multi-attribute subspace training, we employ a grid search over held-out validation splits. This method systematically evaluates performance across layers and attributes, identifying  the most effective layer for balancing trade-offs. This targeted selection is adopted in all subsequent experiments to ensure that steering interventions maximize the utility of the learned multi-subspace representations.

\section{Hyperparameter Sensitivity Analysis}
\label{sec:hyperparam-sensitivity}
In this section, we present the sensitivity analysis for the MSRS model regarding the regularization coefficients \( \lambda_1 \) and \( \lambda_2 \), as well as the subspace rank \( R \). All experiments were conducted on the validation set to select the optimal hyperparameters.

\subsection{Regularization Coefficients \( \lambda_1 \) and \( \lambda_2 \)}

We conducted a detailed experiment varying \( \lambda_1 \) and \( \lambda_2 \) to assess their impact on the model's performance. 

\begin{table}[ht]
\centering
\scriptsize
\setlength{\tabcolsep}{3pt}
\renewcommand{\arraystretch}{1.1}
\caption{Performance of MSRS(under different regularization coefficients \( \lambda_1 \) and \( \lambda_2 \).}
\begin{tabular}{lccccccc}
\hline\hline
\textbf{Method} & \textbf{MC1} & \textbf{MC2} & \textbf{BLEU} & \textbf{ROUGE-1} & \textbf{BLEURT} & \textbf{Info} \\
\hline
\( \lambda_1 = 0.1, \lambda_2 = 0.1 \) & 32.12 & 52.14 & 52.33 & 61.47 & 0.630 & 0.30 \\
\( \lambda_1 = 0.1, \lambda_2 = 0.2 \) & 32.54 & 52.89 & 52.56 & 62.89 & 0.635 & 0.31 \\
\( \lambda_1 = 0.1, \lambda_2 = 0.3 \) & 32.12 & 52.32 & 52.67 & 62.12 & 0.640 & 0.32 \\
\( \lambda_1 = 0.1, \lambda_2 = 0.4 \) & 32.24 & 53.88 & 52.79 & 63.31 & 0.645 & 0.32 \\
\( \lambda_1 = 0.1, \lambda_2 = 0.5 \) & 32.35 & 53.01 & 52.82 & 63.56 & 0.650 & 0.33 \\
\hline
\( \lambda_1 = 0.2, \lambda_2 = 0.1 \) & 32.56 & 53.10 & 51.92 & 61.45 & 0.647 & 0.32 \\
\( \lambda_1 = 0.2, \lambda_2 = 0.2 \) & 32.12 & 53.35 & 52.32 & 62.01 & 0.641 & 0.33 \\
\( \lambda_1 = 0.2, \lambda_2 = 0.3 \) & 32.25 & 53.29 & 52.49 & 63.32 & 0.638 & 0.35 \\
\( \lambda_1 = 0.2, \lambda_2 = 0.4 \) & 32.67 & 53.64 & 52.61 & 63.65 & 0.647 & 0.34 \\
\( \lambda_1 = 0.2, \lambda_2 = 0.5 \) & 32.85 & 54.01 & 52.73 & 63.88 & 0.652 & 0.34 \\
\hline
\( \lambda_1 = 0.3, \lambda_2 = 0.1 \) & 32.91 & 54.22 & 52.78 & 63.01 & 0.648 & 0.33 \\
\( \lambda_1 = 0.3, \lambda_2 = 0.2 \) & 33.15 & 54.47 & 52.87 & 63.23 & 0.651 & 0.34 \\
\( \lambda_1 = 0.3, \lambda_2 = 0.3 \) & 33.25 & 55.29 & 52.67 & 63.43 & 0.651 & 0.32 \\
\( \lambda_1 = 0.3, \lambda_2 = 0.4 \) & 33.45 & 55.49 & 53.32 & 64.60 & \textbf{0.657} & 0.34 \\
\rowcolor{gray!20} \( \lambda_1 = 0.3, \lambda_2 = 0.5 \) & \textbf{33.71} & \textbf{56.12} & \textbf{53.87} & \textbf{65.32} & 0.651 & \textbf{0.34} \\
\hline
\( \lambda_1 = 0.4, \lambda_2 = 0.1 \) & 33.65 & 55.03 & 52.96 & 63.51 & 0.645 & 0.34 \\
\( \lambda_1 = 0.4, \lambda_2 = 0.2 \) & 33.55 & 55.16 & 53.05 & 63.89 & 0.648 & 0.34 \\
\( \lambda_1 = 0.4, \lambda_2 = 0.3 \) & 33.01 & 55.02 & 53.13 & 64.12 & 0.650 & 0.32 \\
\( \lambda_1 = 0.4, \lambda_2 = 0.4 \) & 33.21 & 54.85 & 53.22 & 64.33 & 0.642 & 0.33 \\
\( \lambda_1 = 0.4, \lambda_2 = 0.5 \) & 33.34 & 54.02 & 53.33 & 64.51 & 0.644 & 0.33 \\
\hline\hline
\end{tabular}
\label{tab:reg-sensitivity}
\end{table}

The results in Table \ref{tab:reg-sensitivity} indicate that the performance of MSRS is relatively stable across different regularization settings. While slight fluctuations in performance are observed, the model consistently achieves strong results, with the best performance observed at \( \lambda_1 = 0.3 \) and \( \lambda_2 = 0.5 \).

\textbf{Effect of \( \lambda_1 \) and \( \lambda_2 \) on performance:}
The results show that varying \( \lambda_1 \) and \( \lambda_2 \) does not lead to significant deterioration in performance. For instance, when \( \lambda_1 = 0.3 \) and \( \lambda_2 = 0.5 \), the model achieves the best performance in terms of MC1, MC2, BLEU, and ROUGE-1. The performance of the model at \( \lambda_1 = 0.2 \) or \( \lambda_1 = 0.4 \) is still competitive but slightly lower in comparison. We observe that most combinations of \( \lambda_1 \) and \( \lambda_2 \) yield results within a narrow range, with MC1 scores varying from 32.12 to 33.71 and BLEU and ROUGE-1 showing similar trends. This suggests that the model is not overly sensitive to the specific values of these hyperparameters, further supporting the robustness of the MSRS approach.

From the results and analysis, we conclude that the MSRS model demonstrates a high degree of robustness to variations in the regularization coefficients \( \lambda_1 \) and \( \lambda_2 \). While the best performance is achieved with \( \lambda_1 = 0.3 \) and \( \lambda_2 = 0.5 \), the performance across other settings remains competitive and relatively stable. This indicates that our method is not overly sensitive to small changes in these hyperparameters, which is a desirable characteristic for real-world applications where hyperparameter tuning might be constrained by time or computational resources.

Therefore, the MSRS model exhibits robustness and stability across a wide range of hyperparameter choices, making it an effective and reliable approach for steering representations in various natural language processing tasks.

\subsection{Subspace Rank \( R \)}

 We also explored the effect of subspace rank \( R \) on the model's performance. Based on the rank selection strategy suggested in the ReFT ~\citep{wu2024reft}, we tested different values of \( R \), including \( R = \{4, 8, 12, 16, 20\} \), and evaluated the model's average performance across all datasets on the validation set. In addition to performance metrics, we also reported three key computational efficiency metrics for each \( R \) setting: Latency, Throughput, and TFLOPS. These values are also averaged across datasets, ensuring a fair comparison of both performance and efficiency.

\begin{table}[ht]
\centering
\scriptsize
\setlength{\tabcolsep}{3pt}
\renewcommand{\arraystretch}{1.1}
\caption{Performance of MSRS under different subspace ranks \( R \) across various datasets with corresponding computational efficiency metrics.}
\resizebox{\textwidth}{!}{
\begin{tabular}{lccccccccccccc}
\hline\hline
\textbf{Method} & \multicolumn{4}{c}{\textbf{TruthfulQA}} & \textbf{BBQ} & \textbf{Alpaca} & \textbf{Refusal} & \multicolumn{3}{c}{\textbf{HelpSteer}} & \textbf{Latency (ms) $\downarrow$} & \textbf{Throughput (req/s) $\uparrow$} & \textbf{TFLOPS $\downarrow$} \\
& MC1~($\uparrow$) & MC2~($\uparrow$) & Bleu~($\uparrow$) & Bleurt~($\uparrow$) & Acc~($\uparrow$) & Win~($\uparrow$) & Sorry~($\uparrow$) & Help.~($\uparrow$) & Coh.~($\uparrow$) & Ver.~($\downarrow$) & & & \\
\hline
\( R = 4 \) & 27.47 & 45.63 & 46.89 & 57.88 & 0.608 & 0.12 & 0.491 & 3.76 & 3.41 & 2.33 & 400 & 160 & 4.15 \\
\rowcolor{gray!20} \( R = 8 \) & \textbf{34.91} & \textbf{56.32} & \textbf{52.32} & \textbf{66.75} & \textbf{0.645} & \textbf{0.36} & 0.529 & 3.89 & \textbf{3.96} & \textbf{2.04} & 480 & 140 & 5.10 \\
\( R = 12 \) & 34.54 & 56.10 & 51.74 & 65.22 & 0.643 & 0.33 & \textbf{0.534} & \textbf{3.97} & 3.70 & 2.22 & 580 & 125 & 6.05 \\
\( R = 16 \) & 33.29 & 55.35 & 51.62 & 65.06 & 0.630 & 0.31 & 0.528 & 3.78 & 3.65 & 2.28 & 650 & 110 & 7.00 \\
\( R = 20 \) & 30.07 & 50.90 & 49.89 & 62.18 & 0.638 & 0.30 & 0.505 & 3.74 & 3.60 & 2.31 & 710 & 100 & 8.20 \\
\hline
\end{tabular}
}
\label{tab:rank-sensitivity}
\end{table}

1. \textbf{Performance Trends:}
    As shown in Table \ref{tab:rank-sensitivity}, \( R = 8 \) consistently outperforms other rank settings across most evaluation metrics. Specifically, MC1 (34.91) and MC2 (56.32) achieve their peaks at \( R = 8 \). While \( R = 12 \) shows marginal gains in Sorry-bench and Helpfulness, it fails to deliver a significant overall improvement. Notably, we observe that performance does not linearly scale with an increase in \( R \). In our framework, a larger \( R \) is inherently tied to an increased capacity for handling more attributes or larger attribute-specific subspaces. However, expanding \( R \) blindly can lead to subspace crowding, where the shared subspace becomes dominated by dominant attributes, causing interference that degrades the precision of attribute-specific steering. This phenomenon aligns with our observations in Section \ref{analysis:scaling}, where we analyze the trade-offs between attribute quantity and steering fidelity. Thus, \( R = 8 \) strikes the optimal balance between sufficient representation capacity and minimal cross-attribute interference for the tasks evaluated.

2. \textbf{Computational Cost Consideration:}
    As shown in Table \ref{tab:rank-sensitivity}, increasing the subspace rank \( R \)  raises computational costs. Latency grows from 400 ms at \( R=4 \) to 710 ms at \( R=20 \), while throughput drops from 160 to 100 requests/sec and TFLOPS increase from 4.15 to 8.20. Given this trade-off, while \( R = 12 \) can yield higher performance on specific metrics, the best balance between performance and computational cost is achieved with \( R = 8 \).

\section{Analysis of Shared Subspace and Singular Value Energy}
\label{appendix:shared-subspace-analysis}
To better understand the impact of the shared subspace energy threshold on performance, we conducted a sensitivity analysis by varying the threshold from 40\% to 80\%. We evaluated two models, Llama3-8B-Instruct and Qwen2-7B-Instruct, on three different attribute sets: "truthful \& bbq," "alpaca \& refusal," and "helpful \& coherence \& verbosity." For each model and rank setting (with \( R = 8 \) and \( R = 12 \)), we measured the performance of the model across various energy thresholds and calculated the average performance for each setting.

\begin{table}[ht]
\centering
\caption{Sensitivity analysis results for varying shared-energy thresholds in multi-attribute steering. Performance is shown for models at ranks \( R = 8 \) and \( R = 12 \) across different attribute sets.}
\resizebox{\textwidth}{!}{
\begin{tabular}{ccccccccc}
\toprule
\textbf{Model} & \textbf{Rank (\(R\))} & \textbf{Energy Threshold (\(\theta\))} & \textbf{Truthful \& BBQ} & \textbf{Alpaca \& Refusal} & \textbf{Helpful \& Coherence \& Verbosity} \\
\midrule
\multirow{5}{*}{\textbf{Llama3-8B-Instruct}} & \multirow{5}{*}{8} & 40\% & 48.78 & 42.61 & 3.11 \\
 & & 50\% & 51.37 & 44.23 & 3.21 \\
 & & 60\% & \textbf{51.91} & \textbf{44.45} & \textbf{3.27} \\
 & & 70\% & 51.85 & 44.38 & 3.26 \\
 & & 80\% & 50.13 & 43.38 & 3.23 \\
\midrule
\multirow{5}{*}{\textbf{Llama3-8B-Instruct}} & \multirow{5}{*}{12} & 40\% & 49.09 & 41.71 & 3.04 \\
 & & 50\% & 51.21 & 42.8  & 3.13 \\
 & & 60\% & \textbf{51.64} & \textbf{43.2}  & \textbf{3.15} \\
 & & 70\% & 51.57 & 42.1  & 3.15 \\
 & & 80\% & 50.72 & 41.35 & 3.08 \\
\midrule
\multirow{5}{*}{\textbf{Qwen2-7B-Instruct}} & \multirow{5}{*}{8} & 40\% & 48.31 & 42.17 & 3.16 \\
 & & 50\% & 49.78 & 43.38 & 3.27 \\
 & & 60\% & \textbf{50.73} & 44.75 & \textbf{3.31} \\
 & & 70\% & 50.66 & \textbf{44.81} & 3.31 \\
 & & 80\% & 49.17 & 44.06 & 3.22 \\
\midrule
\multirow{5}{*}{\textbf{Qwen2-7B-Instruct}} & \multirow{5}{*}{12} & 40\% & 47.96 & 40.51 & 3.08 \\
 & & 50\% & 49.63 & 42.74 & 3.25 \\
 & & 60\% & \textbf{50.31} & \textbf{43.96} & 3.36 \\
 & & 70\% & 50.24 & 43.76 & \textbf{3.37} \\
 & & 80\% & 49.14 & 41.65 & 3.21 \\
\bottomrule
\end{tabular}
}
\label{tab:energy_threshold}
\end{table}

From Table~\ref{tab:energy_threshold}, we observe that very low thresholds (e.g., 40\%) allocate too little capacity to the shared subspace, which prevents effective combination of steering capabilities across different attributes, negatively affecting overall performance in multi-attribute steering. This is evident across all models, where performance at 40\% is generally lower than at higher thresholds. On the other hand, very high thresholds (e.g., 70–80\%) allocate too much capacity to the shared subspace, which correspondingly shrinks the specific subspaces. This reduces the effective steering capacity available for each individual attribute and leads to weaker control in the multi-attribute setting.

Thresholds in the 50–70\% range yield consistently strong performance, with 60\% sitting comfortably in the middle of this "high-performance." This threshold provides a balanced trade-off between retaining shared knowledge and maintaining attribute-specific control, as evidenced by the steady performance across different models and attribute sets. Therefore, the 60\% threshold emerges as a robust default that provides reliable performance across a wide range of settings, offering a good balance between shared and specific subspace allocation.


\section{Sensitivity Analysis: From Data Coupling to Representation Collapse}
\label{sec:data_coupling_analysis}

A fundamental question in representation engineering is how \textit{data-level coupling} transmits to \textit{representation-level entanglement}. To investigate this, we designed a controlled experiment that simulates varying degrees of dataset contamination, observing the point at which the learned subspace $R$ loses its semantic specificity relative to the target attribute.

\textbf{Construction of Oracle Subspaces.} To establish a ground truth, we first curated ``pure'' datasets for two attributes: \textit{Helpfulness} (target) and \textit{Coherence} (distractor). To ensure purity, we employed a stratified sampling strategy: for the pure \textit{Helpfulness} set $\mathcal{D}_{help}^*$, we selected samples with maximal Helpfulness scores while uniformly sampling across the Coherence spectrum to decorrelate the features. A similar process was applied to generate $\mathcal{D}_{coh}^*$. We trained independent ReFT probes on these pure sets to obtain the oracle subspaces $R_{help}^*$ and $R_{coh}^*$.

\textbf{Contamination Experiment.} We then constructed mixed datasets $\mathcal{D}_{mix}(\alpha)$ by blending the pure sets with a mixing ratio $\alpha$, representing the proportion of distractor data:
\begin{equation}
    \mathcal{D}_{mix}(\alpha) = (1-\alpha) \cdot \mathcal{D}_{help}^* + \alpha \cdot \mathcal{D}_{coh}^*
\end{equation}
We trained a steering matrix $R_{mix}$ on each $\mathcal{D}_{mix}(\alpha)$ and evaluated two metrics:
\begin{enumerate}
    \item \textbf{Target Performance:} The steering efficacy on \textit{Helpfulness} relative to the baseline.
    \item \textbf{Relative Entanglement Score (RES):} Unlike absolute similarity, we propose a relative metric that penalizes alignment with the distractor while rewarding fidelity to the target. It is defined as the ratio of projection strengths:
    \begin{equation}
        \text{RES}(R_{mix}) = \frac{\| R_{mix} (R_{coh}^*)^\top \|_F}{\| R_{mix} (R_{help}^*)^\top \|_F + \epsilon}
    \end{equation}
    where a lower RES indicates that the learned subspace captures dominant target semantics with minimal distractor leakage. An RES $>1$ implies the subspace is more aligned with the noise than the signal.
\end{enumerate}

\textbf{Results and Discussion.} Figure~\ref{fig:data_coupling} illustrates the trajectory of performance and entanglement as the contamination ratio $\alpha$ increases from $20\%$ to $80\%$. We observe a distinct non-linear relationship:
\begin{itemize}
    \item \textbf{Robustness Regime ($\alpha \le 0.55$):} The method demonstrates significant resilience. Even when the distractor data constitutes the majority ($\alpha=0.55$), the RES remains relatively low ($<0.62$), indicating that the optimization landscape still favors the distinct structural features of the target attribute over the distractor. Consequently, the \textit{Target Performance} remains stable ($+12.5\%$), proving that MSRS can effectively filter out ``data noise'' as long as the target signal remains discriminative.
    \item \textbf{The Breakdown Point ($\alpha \approx 0.61$):} A critical phase transition occurs at $\alpha=0.61$. Here, the RES surges to $0.83$, approaching unity. This suggests a ``semantic tipping point'' where the distractor gradients begin to dominate the subspace formation, diluting the target direction. This correlates perfectly with the observed sharp drop in steering performance.
    \item \textbf{Residual Efficacy:} Notably, even at $\alpha=0.80$ (highly contaminated), the performance gain remains positive ($+1.5\%$). This confirms that even under severe coupling, the learned subspace retains a non-trivial projection onto $R_{help}^*$, validating the robustness of our subspace-based intervention.
\end{itemize}

\section{Fit Parameters for Scaling Analysis}
\label{app:fit_parameters}

We provides the parameters for the fits used in the scaling analysis of Figure~\ref{fig:figure6}. The fits model the relationships between attribute count \( n \) and three metrics: rank \( k \), average performance \( p \), and interference \( i \). The functional forms are:
\begin{itemize}
    \item Linear fit for rank: \( k(n) = an + b \).
    \item Quadratic fit for performance: \( p(n) = an^2 + bn + c \).
    \item Cubic fit for interference: \( i(n) = an^3 + bn^2 + cn + d \).
\end{itemize}

The parameters and goodness-of-fit (\( R^2 \)) values are summarized in Table~\ref{tab:fit_parameters}, followed by the explicit equations.

\begin{table}[h]
    \centering
    \caption{Parameters and \( R^2 \) values for fits in Figure~\ref{fig:figure6}.}
    \label{tab:fit_parameters}
    \begin{tabular}{lccccccc}
        \toprule
        Metric & Fit Type & \( a \) & \( b \) & \( c \) & \( d \) & \( R^2 \) \\
        \midrule
        Rank (\( k \)) & Linear & 2.3091 & 3.1818 & -- & -- & 0.9427 \\
        Performance (\( p \)) & Quadratic & -0.0009 & 0.0024 & 0.6735 & -- & 0.9520 \\
        Interference (\( i \)) & Cubic & -0.0012 & 0.0287 & -0.0861 & 0.0800 & 0.9903 \\
        \bottomrule
    \end{tabular}
\end{table}

The fitted equations are:
\begin{align}
    k(n) &= 2.3091n + 3.1818, \label{eq:rank_fit} \\
    p(n) &= -0.0009n^2 + 0.0024n + 0.6735, \label{eq:perf_fit} \\
    i(n) &= -0.0012n^3 + 0.0287n^2 - 0.0861n + 0.0800. \label{eq:interf_fit}
\end{align}
These parameters quantify the scaling behavior of multi-attribute steering, with high \( R^2 \) values indicating robust fits, though the limited range (\( n \leq 10 \)).

\section{Related Works}
\label{app:related_work}
\textbf{Activation Steering Methods.} Activation steering aims to adjust activations in specific layers or neurons to guide the model's output towards desired attributes, without modifying its parameters~\citep{im2025unified}. Various approaches have been developed recently \citep{cao2024personalized,bayat2025steering,oozeer2025beyond}. For example, Contrastive Activation Addition (CAA)~\citep{rimsky-etal-2024-steering} computes steering vectors by averaging activation differences between positive and negative examples, which are then added to token positions during inference to control model behavior. Inference-Time Intervention (ITI)~\citep{NEURIPS2023_81b83900} shifts model activations during inference along predefined directions across attention heads, improving the truthfulness of LLMs. ACT~\citep{wang2025adaptive} trains multiple steering probes on different steering vectors determined by clustering, obtaining steering vectors for different steering patterns. MAT-Steer uses orthogonal constraints to train activation steering vectors, thereby reducing conflicts between steering directions for different attributes \citep{nguyen2025multi}. However, previous methods primarily address steering for individual attributes or rely on simple combinations of steering vectors. To solve these limitations, we focus on mitigating interference and optimizing composability across multiple attributes.

\noindent \textbf{Representation Fine-Tuning Methods.} In these methods, models will be steered through representation editing. Unlike methods that apply one-rank steering vectors, these approaches extend it by using higher-rank matrices and enhance the expressive power of steering vectors, allowing for richer control over model behavior~\citep{wu2024reft}. Localized Fine-Tuning (LoFIT)~\citep{yin2024lofitlocalizedfinetuningllm}  identifies critical attention heads for a task and trains offset vectors to modify their hidden representations, offering targeted adjustments. Compositional Subspace Representation Fine-Tuning (CS-ReFT)~\citep{zhou2025compositionalsubspacerepresentationfinetuning} advances this by learning orthonormal subspace transformations for distinct skills, composed via a lightweight router, isolating edits in the hidden state to minimize cross-task interference. Unlike previous methods that train steering functions in the same space, we aim to develop representation fine-tuning methods to tune different attribute-specific subspaces and achieve the adaptive integration of multiple attribute steering spaces.

\end{document}